\documentclass[pdflatex,sn-mathphys-num]{sn-jnl}% Math and Physical Sciences Numbered Reference Style
\usepackage{graphicx}%
\usepackage{multirow}%
\usepackage{amsmath,amssymb,amsfonts}%
\usepackage{amsthm}%
\usepackage{mathrsfs}%
\usepackage[title]{appendix}%
\usepackage{xcolor}%
\usepackage{textcomp}%
\usepackage{manyfoot}%
\usepackage{booktabs}%
\usepackage{algorithm}%
\usepackage{algorithmicx}%
\usepackage{algpseudocode}%
\usepackage{listings}%
\usepackage{graphicx}
\usepackage{subcaption}
\usepackage{placeins} 
\usepackage{booktabs}
\usepackage{array} % for >{\centering\arraybackslash}

\theoremstyle{thmstyleone}%
\theoremstyle{thmstyletwo}%

\theoremstyle{thmstylethree}%

\begin{document}

\title[Article Title]{UniCoMTE: A Universal Counterfactual Framework for Explaining Time-Series Classifiers on ECG Data}


\author[1]{\fnm{Justin} \sur{Li}}\email{justinli@bu.edu}

\author[1]{\fnm{Efe} \sur{Sencan}}\email{esencan@bu.edu}

\author[2]{\fnm{Jasper Zheng} \sur{Duan}}\email{jzduan@sandia.gov}

\author[2]{\fnm{Vitus J.} \sur{Leung}}\email{vjleung@sandia.gov}

\author[1,3]{\fnm{Stephen} \sur{Tsaur}}\email{Stephen.Tsaur@bmc.org}

\author[1]{\fnm{Ayse K.} \sur{Coskun}}\email{acoskun@bu.edu}

\affil*[1]{\orgname{Boston University}, \city{Boston},\state{MA}, \country{USA}}

\affil[2]{\orgname{Sandia National Laboratories}, \orgaddress{\city{Albuquerque}, \state{NM}, \country{USA}}}

\affil[3]{\orgname{Boston Medical Center}, \city{Boston}, \state{MA}, \country{USA}}


\abstract{Machine learning models, particularly deep neural networks, have demonstrated strong performance in classifying complex time series data. However, their black-box nature limits trust and adoption, especially in high-stakes domains such as healthcare. To address this challenge, we introduce UniCoMTE, a model-agnostic framework for generating counterfactual explanations for multivariate time series classifiers. The framework identifies temporal features that most heavily influence a model's prediction by modifying the input sample and assessing its impact on the model’s prediction. UniCoMTE is compatible with a wide range of model architectures and operates directly on raw time series inputs. In this study, we evaluate UniCoMTE’s explanations on a time series ECG classifier. We quantify explanation quality by comparing our explanations’ comprehensibility to comprehensibility of established techniques (LIME and SHAP) and assessing their generalizability to similar samples. Furthermore, clinical utility is assessed through a questionnaire completed by medical experts who review counterfactual explanations presented alongside original ECG samples. Results show that our approach produces concise, stable, and human-aligned explanations that outperform existing methods in both clarity and applicability. By linking model predictions to meaningful signal patterns, the framework advances the interpretability of deep learning models for real-world time series applications.
}




\maketitle
\section{Introduction}\label{sec1}
Cardiovascular diseases (CVDs) remain the leading cause of death globally, accounting for an estimated 17.9 million deaths each year~\cite{who-cvd}. Early detection and diagnosis are critical for reducing morbidity and mortality, as timely interventions can significantly improve  outcomes~\cite{almansouri2024early}. Electrocardiograms (ECGs) serve as a primary non-invasive diagnostic tool to assess cardiac function by recording the heart’s electrical activity over time. 
Given the complexity and sheer volume of ECG recordings, researchers have increasingly turned to deep learning methods as a means to automate ECG-based diagnosis. 

Recent studies have demonstrated that deep learning models in particular can achieve high performance for ECG classification tasks and show potential for clinical application in research settings. For example, a deep neural network trained on 12-lead ECG samples can outperform cardiology residents in detecting multiple arrhythmias, with F1-scores above 80\% and specificity over 99\%, across six ECG abnormalities~\cite{ribeiro2020automatic}. Similarly, a Convolutional Neural Network~\cite{o2015introduction} (CNN) model trained on 12-lead ECG data can perform on par with cardiologists and exhibits greater accuracy than a leading commercial ECG analysis system. Other models have achieved high performances across a range of similar classification tasks including the classification of myocardial infarction and atrial fibrillation~\cite{hannun2019cardiologist,attia2019artificial,rajpurkar2017cardiologist}. Beyond performance comparisons with clinical standards, several studies investigate the impact of architectural choices. For instance, using one-dimensional time-series models appear more effective than transforming ECG signals into image representations. One study finds that a gated recurrent unit–based recurrent neural network~\cite{shiri2023comprehensive} achieves around 80\% sensitivity and 81\% specificity, outperforming both two-dimensional CNN approaches and multimodal fusion of one- and two-dimensional inputs.
In terms of efficiency, a lightweight 11-layer hybrid convolutional neural network–long short-term memory (CNN–LSTM) model achieves near-perfect arrhythmia classification (approximately 98\% accuracy) across eight rhythm classes~\cite{alamatsaz2022lightweight}, while remaining compact enough for deployment to wearable monitors for continuous, real-time detection. Traditional feature-based ML methods also show promise: one approach combines advanced ECG signal processing—such as peak detection—with a ML classifier to achieve state-of-the-art heartbeat classification performance on a large dataset of over 10{,}000 patients~\cite{aziz2021ecg}. Notably, this method maintains high accuracy across different patient cohorts, achieving around 80–90\% accuracy even when evaluated on external hospital data, in contrast to sharp performance drops observed in less generalizable models.

Although these models have achieved high performance across a range of disease classification tasks in research settings, clinical integration remains limited. In other words, accurate classification alone is insufficient for clinical adoption of ML models. Black-box predictions without transparent reasoning can undermine clinician trust and patient safety~\cite{pumplun2021adoption, marey2024explainability, quinn2022three}. As a result, a growing body of research focuses on explaining ECG classification models. Recent studies apply various explainable artificial intelligence (XAI) techniques to identify segments of the ECG signal that most heavily influence a model’s prediction. Perturbation-based methods systematically alter segments of the ECG signal to determine the points that are most influential towards the model’s output ~\cite{perturbation2022}. Two widely used techniques include Shapley Additive Explanations~\cite{lundberg2017unified} (SHAP) and Local Interpretable Model-Agnostic Explanations~\cite{ribeiro2016should} (LIME), which approximate the local behavior of a complex model using simpler surrogate models and distribute attribution scores across input features to quantify their influence on the prediction. These methods are often used to establish the relative importance of specific time points or waveform intervals in ECG-based classifications~\cite{shap_hrv, singh2022interpretation, aggarwal2022ecg, sathi2024interpretable}. 

Despite their popularity, SHAP and LIME face important limitations in the clinical context. First, they typically assume feature independence and lack temporal awareness, which make them less reliable when applied to time series data like ECG signals. Second, the resulting explanations, such as attribution maps or abstract score rankings, can be difficult for clinicians to interpret and act upon, especially when they do not clearly display known physiological patterns or the impact of a time series feature on a diagnosis. As such, while current XAI methods offer a starting point for transparency, there remains a critical need for explanation frameworks that produce more intuitive, clinically grounded, and actionable insights~\cite{xai_limitations}. Prior work shows that counterfactual based XAI can enhance clinicians' understanding and trust in imaging-based models~\cite{mertes2022ganterfactual,
SINGLA2023102721}. However, to our knowledge this has yet to be demonstrated with time-series data such as ECGs. Generative counterfactual XAI has recently been developed for ECGs, though this framework may not be readily generalized to other diagnoses or models~\cite{jang2025gcx}.

To address the limitations of traditional explainability methods in time-series classification, CoMTE~\cite{ates2021counterfactual} provides instance-specific counterfactual explanations by identifying minimal changes to the input that flip the model’s prediction. Given a model prediction and a target class of interest, CoMTE searches for a training sample from the target class—referred to as a distractor—and identifies a small set of variables (i.e., time series features) whose substitution with the original sample's corresponding features will causes the model to alter its prediction from the original class to the target class. CoMTE then replaces these variables in the input sample with the corresponding variables from the distractor, generating a counterfactual example that is minimally different from the original input but classified as the target class. This approach helps users understand specific features that contribute most to the model's decision, and how the prediction could change if those features were different. 

Building upon the foundation of CoMTE, we introduce CoMTE-V1.1 (UniCoMTE), which retains the core methodology of generating counterfactual explanations through targeted segment substitution. The primary advancement in UniCoMTE lies in its flexible architecture, which supports a wide range of model backends, including TensorFlow~\cite{tensorflow}, PyTorch~\cite{pytorch}, and scikit-learn~\cite{scikitlearn}. This compatibility allows researchers to use UniCoMTE across different modeling pipelines without modifying the underlying structure of the explainability algorithm. 

We subsequently apply UniCoMTE on a state-of-the-art CNN trained for 12-lead ECG classification. Compared with SHAP and LIME, which output thousands of point-level attribution scores,UniCoMTE produces concise counterfactual explanations involving only 2–4 lead–time segments, directly identifying the minimal changes needed to alter the model's prediction. These explanations are generalizable, successfully altering diseased samples to be classified as normal samples with 43–58\% coverage across six conditions (see Table~\ref{tab:coverage}). Furthermore, responses to a questionnaire completed by clinical experts indicated that, overall, UniCoMTE’s explanations are clinically relevant and easy to interpret. Our contributions can be summarized as follows:

\begin{itemize}
    \item We introduce UniCoMTE, a flexible counterfactual explanation framework that unifies counterfactual reasoning with compatibility across diverse ML libraries and classification algorithms, enabling application to a broader class of time series models.  
    
    \item We apply UniCoMTE to a state-of-the-art CNN trained to classify multivariate ECG signals, enabling counterfactual explanations for clinically relevant cardiac conditions.
    
    \item We demonstrate UniCoMTE's applicability to real-world medical time series data through both quantitative benchmarking—an assessment of comprehensibility and generalizability—and qualitative expert review, where practicing clinicians assess the clinical relevance and interpretability of generated explanations through a structured questionnaire.
    
\end{itemize}

\section{Results}\label{sec2}
\subsection{Quantitative Evaluation of UniCoMTE Explanations}

We evaluate UniCoMTE to assess its ability to generate interpretable, reliable, and clinically meaningful explanations for ECG classification models. Our evaluation follows two complementary approaches: quantitative and qualitative analysis. The quantitative evaluation examines two measurable properties of the generated explanations—\textit{comprehensibility} and \textit{generalizability}—adopted from the evaluation methodology of the original CoMTE framework~\cite{ates2021counterfactual} to ensure consistency and comparability. The qualitative evaluation assesses clinical interpretability through a structured questionnaire with medical experts, providing insight into the practical utility of UniCoMTE’s explanations.

\subsubsection{Comprehensibility}

We evaluate the \textit{comprehensibility} of UniCoMTE, defined as how easily a human user can understand which regions of the ECG drive the model’s prediction and how minimal, localized changes could alter that decision. An explanation is considered more comprehensible when it highlights only a small number of variables that are sufficient to explain the impact of features on a model’s output.

\begin{figure}[!t]
    \centering
    \includegraphics[width=0.6\linewidth]{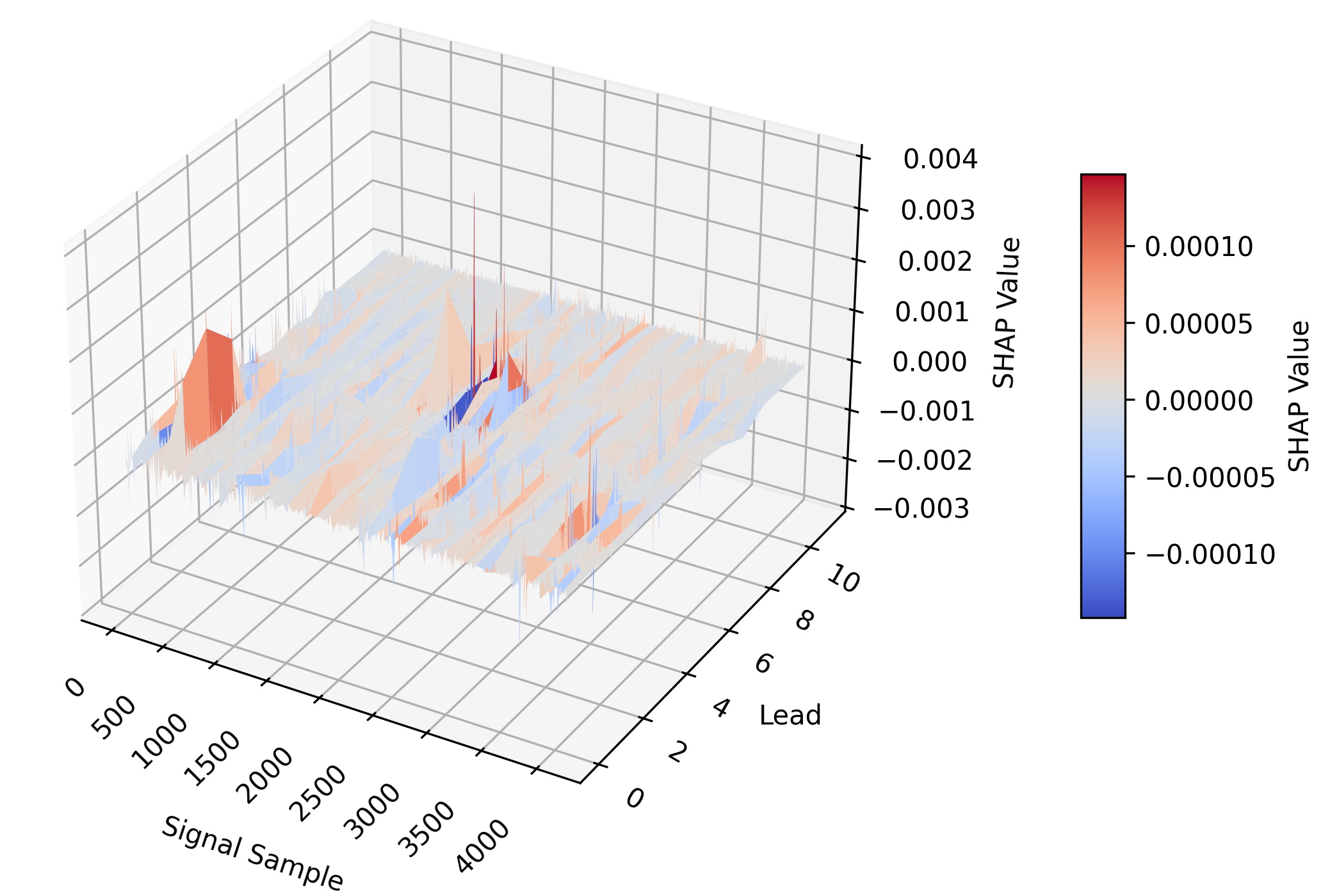}
    \caption{SHAP surface plot for first-degree atrioventricular block (1dAVb). The plot shows how individual ECG samples across 12 leads contribute to the classification decision.}
    \label{fig:shap-surface}
\end{figure}

\begin{figure}[t!]
    \centering
    \includegraphics[width=0.5\linewidth]{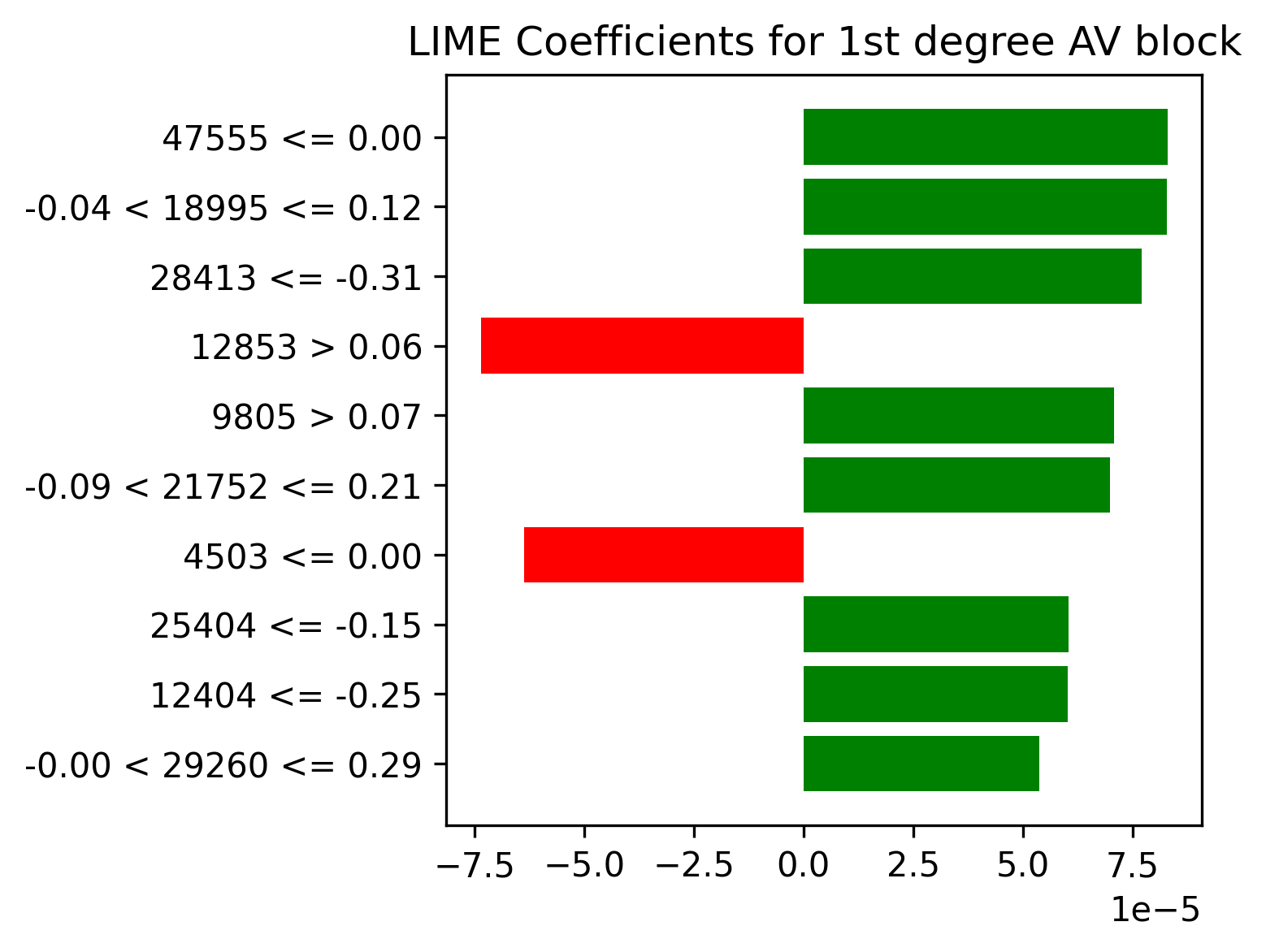}
    \caption{Example LIME output showing the ten most influential features for a misclassified ECG sample. The bar chart illustrates relative importance weights assigned to signal.}
    \label{fig:lime}
    \vspace{-1.5em}
\end{figure}

We compare UniCoMTE with SHAP and LIME, two widely used explainability baselines for ECG classification. SHAP produces a dense tensor of attribution values—one for each time sample and ECG lead—representing how local variations in the input influence the output. For instance, when applied to first-degree atrioventricular block, SHAP coefficients can be represented with a three-dimensional surface plot. The horizontal axes correspond to the time point and ECG lead while the vertical axis indicate the contribution strength of each segment (Fig.~\ref{fig:shap-surface}). Although SHAP highlights signal regions that influence the model’s decision, its fine-grained attributions span thousands of data points, resulting in explanations that are difficult to interpret without manual thresholding or aggregation. Furthermore, it is unclear the extent to which these features influence the model's prediction.  
LIME perturbs each input instance and fits a local linear model to approximate the classifier’s behavior. For ECG data, it returns a ranked list of influential features and their corresponding weights. In Fig.~\ref{fig:lime}, each row corresponds to a discretized feature interval produced by LIME’s  \texttt{TabularExplainer}, and the horizontal axis represents the signed coefficient of that feature in the local linear surrogate model. Green bars indicate features that increase the model’s confidence in the predicted class, while red bars indicate features that suppress it; the bar length reflects the magnitude of influence. The number of displayed features must be predefined by the user; for illustration, we set \texttt{num\_features}=10. While LIME provides a ranked summary of the most influential signal indices, it does not specify the extent in which changes to these regions would alter the model’s prediction, which limits its clinical interpretability.  

We quantify comprehensibility by measuring the number of features required for an explanation—the fewer features, the easier it is for a human reader to interpret the model’s reasoning. Across test samples that are classified as abnormal, UniCoMTE identifies an average of 2.93 lead–time segments that must be modified to flip the model’s prediction. Most samples require 2 lead-time segments to be modified. In contrast, SHAP produces thousands of coefficients and LIME returns an arbitrary number of user-defined features.  
From a technical standpoint, both SHAP and LIME operate on point-level attributions that describe feature importance in a relative rather than absolute sense. They reveal which regions are more influential than others but not how much a given alteration would impact the model’s decision. As a result, users must interpret these explanations indirectly—by thresholding importance scores and mapping them back onto ECG signals—a process that is subjective and often inconsistent across datasets. These challenges become more pronounced for multivariate time series, where dependencies among leads and temporal dynamics cannot be captured by independently weighting features.  

UniCoMTE, in contrast, performs explanation at the level of ECG leads and temporal segments rather than individual points. By identifying the minimal subset of segments whose replacement changes the model’s prediction, the framework provides a direct, actionable view of the decision boundary. Consequently, UniCoMTE delivers concise, human-readable explanations that are both technically faithful to the model’s behavior and semantically aligned with how clinicians interpret ECGs.

\subsubsection{Generalizability}

\begin{table}[t!]
\centering
\caption{Coverage results for different misclassification types. $N$ denotes the total number of misclassified samples tested for each type.}
\label{tab:coverage}
\begin{tabular}{lcc}
\toprule
\textbf{Misclassification Type (True, Predicted)} & \textbf{Coverage (\%)} & \textbf{N} \\
\midrule
Normal, 1dAVb & 57 & 49 \\
Normal, RBBB  & 49 & 34 \\
Normal, LBBB  & 47 & 38 \\
Normal, SB    & 58 & 32 \\
Normal, AF    & 43 & 8  \\
Normal, ST    & 53 & 35 \\
\bottomrule
\end{tabular}
\end{table}

We define \textit{generalizability} as the ability of an explanation generated for one misclassified sample to also correct other misclassifications of the same type. UniCoMTE identifies the minimal set of lead–time segments whose modification changes a misclassified sample to its correct label. We then apply these same segment changes to other samples that share the same incorrect prediction and count each successful correction as a \textit{hit}. We measure generalizability using \textit{coverage}, the ratio of hits to the total number of tested misclassifications of that type. A higher coverage value indicates that a single explanation corrects multiple similar misclassifications rather than only one case.  
Table~\ref{tab:coverage} summarizes the coverage results across six diagnostic conditions. UniCoMTE achieves coverage between 43\% and 58\%, with the highest values for Normal–sinus bradycardia (58\%) and Normal–first-degree atrioventricular block (57\%). Even the lowest coverage, for Normal–atrial fibrillation (43\%), shows that a single explanation corrects a substantial fraction of similar misclassifications. These results show that UniCoMTE identifies recurring patterns in the model’s predictions, where the same signal regions contribute to repeated errors across samples of the same type. By revealing these consistent decision patterns, UniCoMTE helps characterize how the model differentiates normal from abnormal ECGs and highlights areas where it tends to make similar mistakes.

% \subsubsection{Generalizability:} 
% We define generalizability as the ability of an explanation to correct not only the specific misclassification it is generated for but also other misclassifications of the same type. In practice, this means that if UniCoMTE identifies a minimal set of changes that flips a Normal sample misclassified as 1dAVb to its correct label, we then test whether the same changes also correct other Normal samples misclassified as 1dAVb. Each time the explanation successfully flips the prediction, we count it as a \textit{hit}. We measure generalizability using \textit{coverage}, the ratio of hits to the total number of tested misclassifications of that type. A higher coverage value indicates that the explanation captures a recurring misclassification pattern rather than a one-off case. As shown in Table~\ref{tab:coverage}, coverage ranges from 43\% to 58\% across conditions, with the highest values for Normal–SB (58\%) and Normal–1dAVb (57\%). Even the lowest coverage, observed for Normal–AF (43\%), shows that UniCoMTE explanations can generalize to a substantial fraction of recurring errors.

% These promising coverage values are greater than the ones presented in the original CoMTE implementation on high performance computing data~\cite{ates2021counterfactual}. Therefore, these results indicate that UniCoMTE's feature-level time series substitution method can identify underlying trends that are indicative of each cardiovascular condition. 

\subsection{Qualitative Evaluation of Counterfactual Explanations}

\begin{table}[b!]
\centering
\caption{Survey Responses and Likert Scale}
\label{tab:ratings}
\begin{tabular}{%
  >{\centering\arraybackslash}p{0.60\linewidth}
  >{\centering\arraybackslash}p{0.25\linewidth}%
}
\toprule
\textbf{To what extent does the explanation make the diagnostic criteria for the abnormal ECG more apparent?}
&
\textbf{Associated Quantitative Score} \\
\midrule
Very Helpful & 5 \\
Helpful & 4 \\
Neutral & 3 \\
Slightly Misleading / Incomplete Explanation & 2 \\
Misleading & 1 \\
\bottomrule
\end{tabular}
\end{table}

Quantitative metrics capture certain properties of explanations, but they cannot fully assess whether explanations are clinically meaningful. To complement our quantitative results, we conduct a structured questionnaire administered to medical experts, who directly evaluate the quality and utility of UniCoMTE’s counterfactual explanations. Experts are presented with ECG samples and asked to rate how effective the explanations were in highlighting attributes in ECG waveforms that explain the sample's classification (see Table~\ref{tab:ratings}). This evaluation allows us to determine whether the explanations highlight signal changes that clinicians regard as plausible and informative for distinguishing between normal and abnormal ECG samples. In other words, the questionnaire probes the central question: do UniCoMTE's explanations help experts identify attributes that would make an abnormal sample appear normal, thereby providing insight into the model’s behavior? 

\begin{figure}[H]
    \centering
    \includegraphics[width=0.5\linewidth]{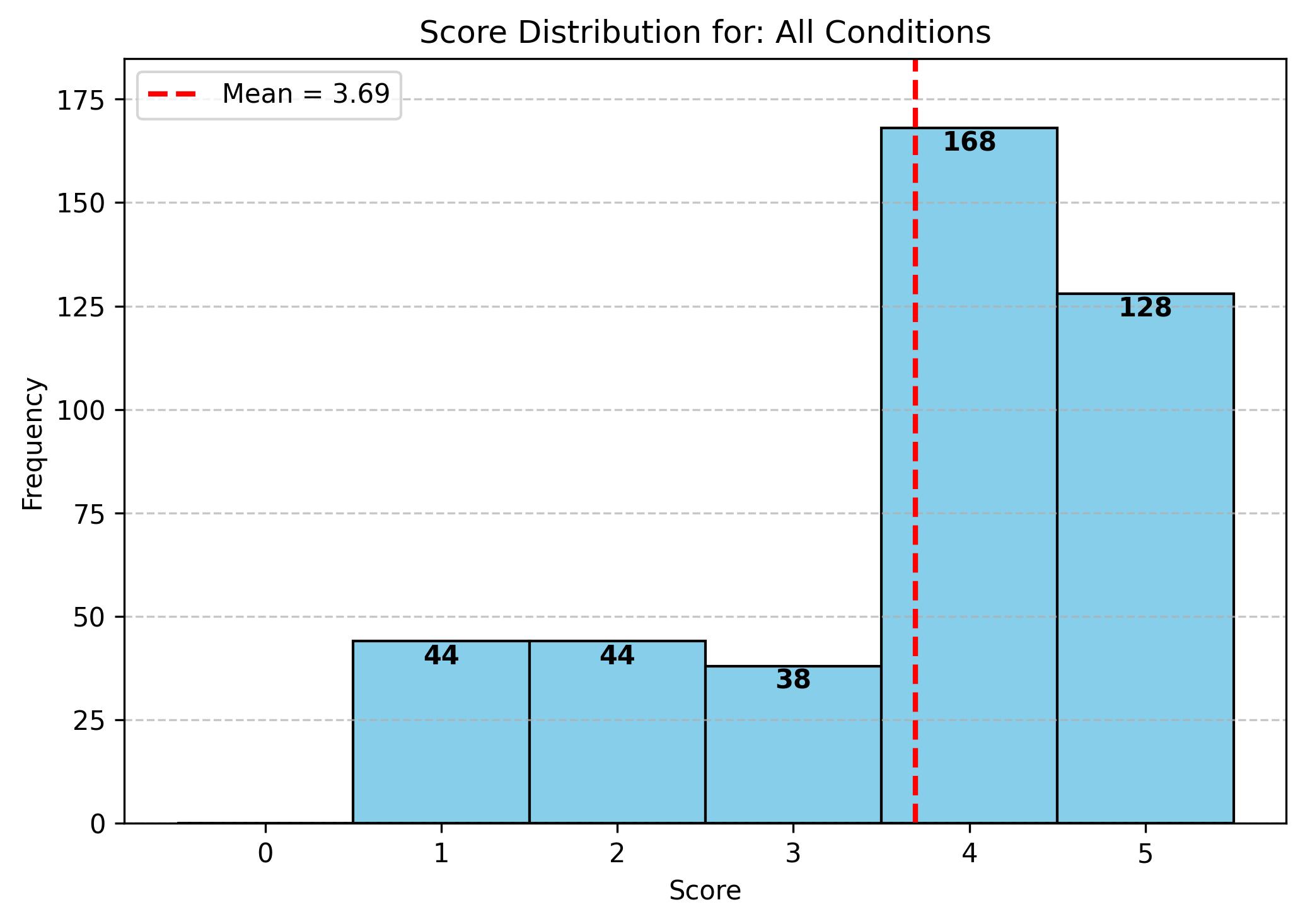}
    \caption{Distribution of scores across all expert responses and conditions}
    \label{fig:all}

    % Row 1
    \begin{subfigure}{0.48\linewidth}
        \includegraphics[width=\linewidth]{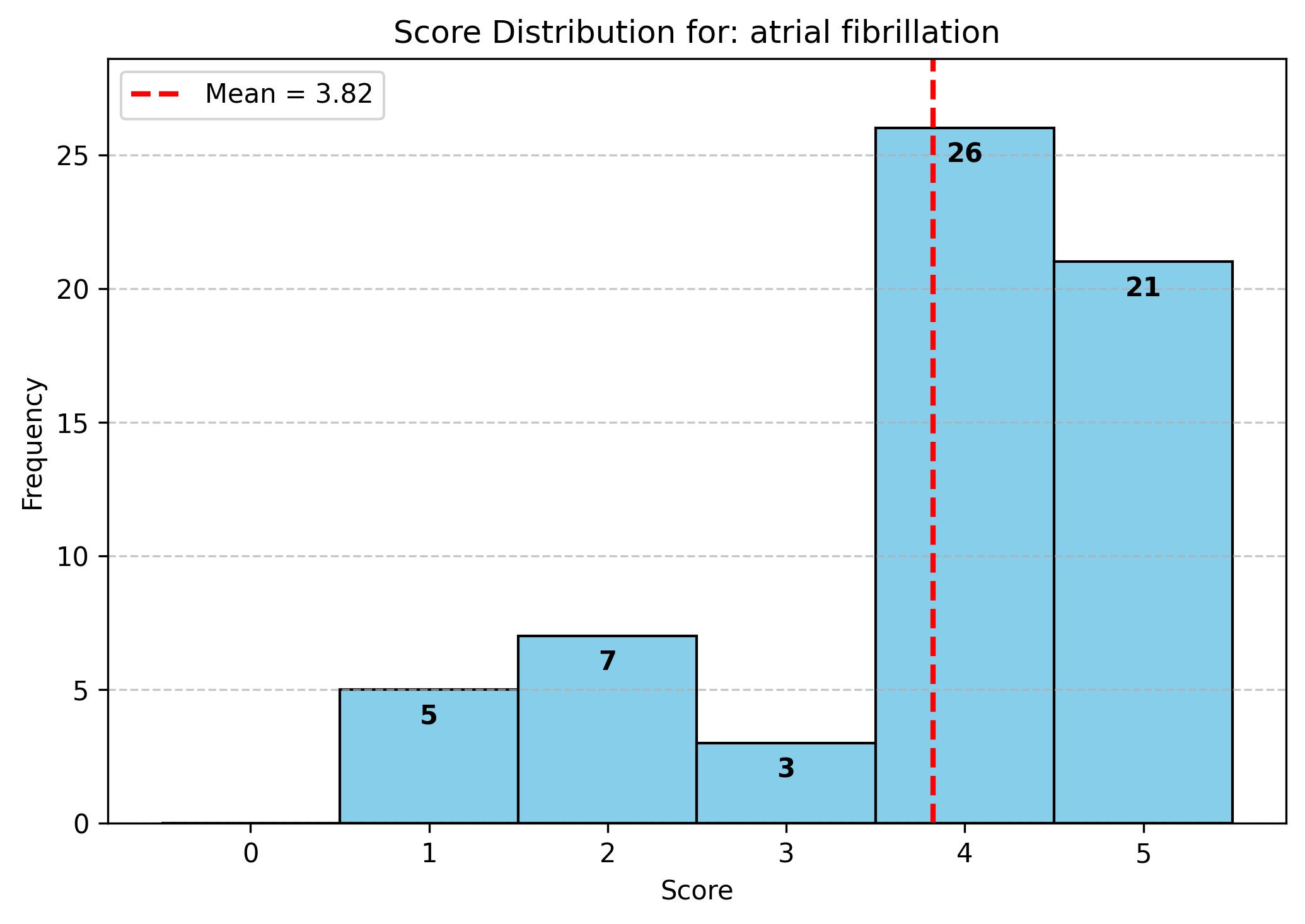}
        \caption{Atrial fibrillation}
        \label{fig:af}
    \end{subfigure}
    \hfill
    \begin{subfigure}{0.48\linewidth}
        \includegraphics[width=\linewidth]{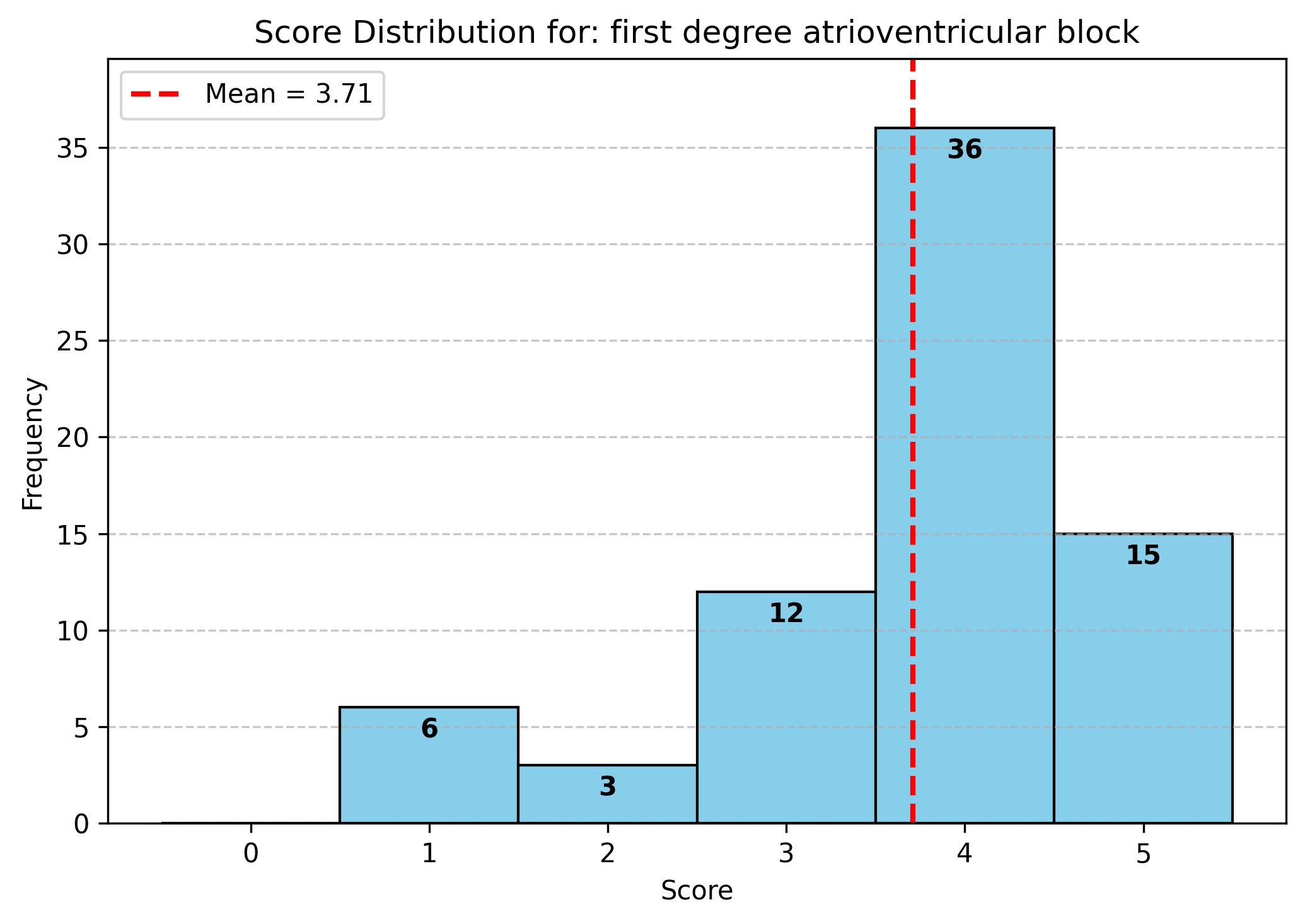}
        \caption{First-degree AV block}
        \label{fig:avblock}
    \end{subfigure}

    % Row 2
    \begin{subfigure}{0.48\linewidth}
        \includegraphics[width=\linewidth]{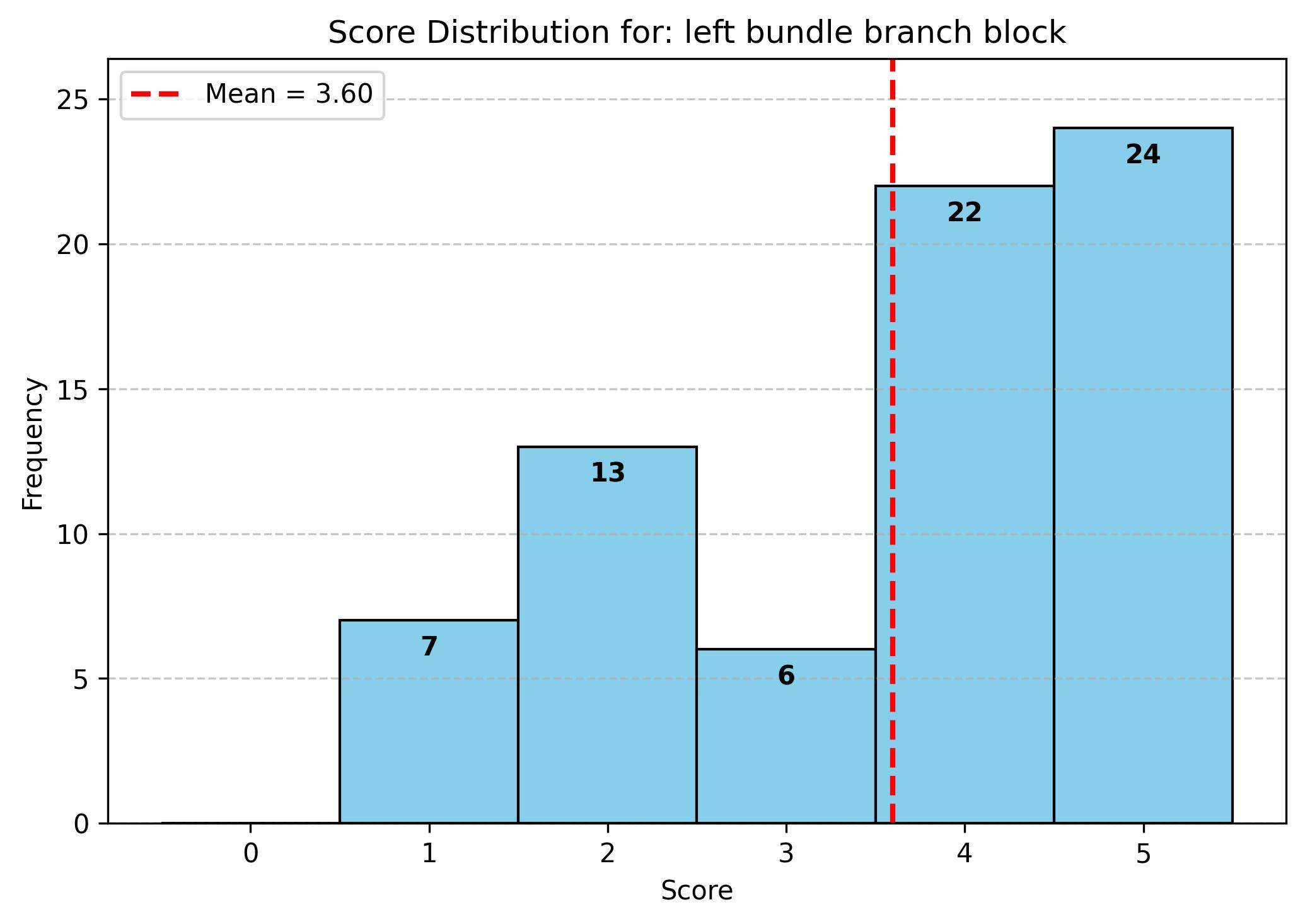}
        \caption{Left bundle branch block}
        \label{fig:lbbb}
    \end{subfigure}
    \hfill
    \begin{subfigure}{0.48\linewidth}
        \includegraphics[width=\linewidth]{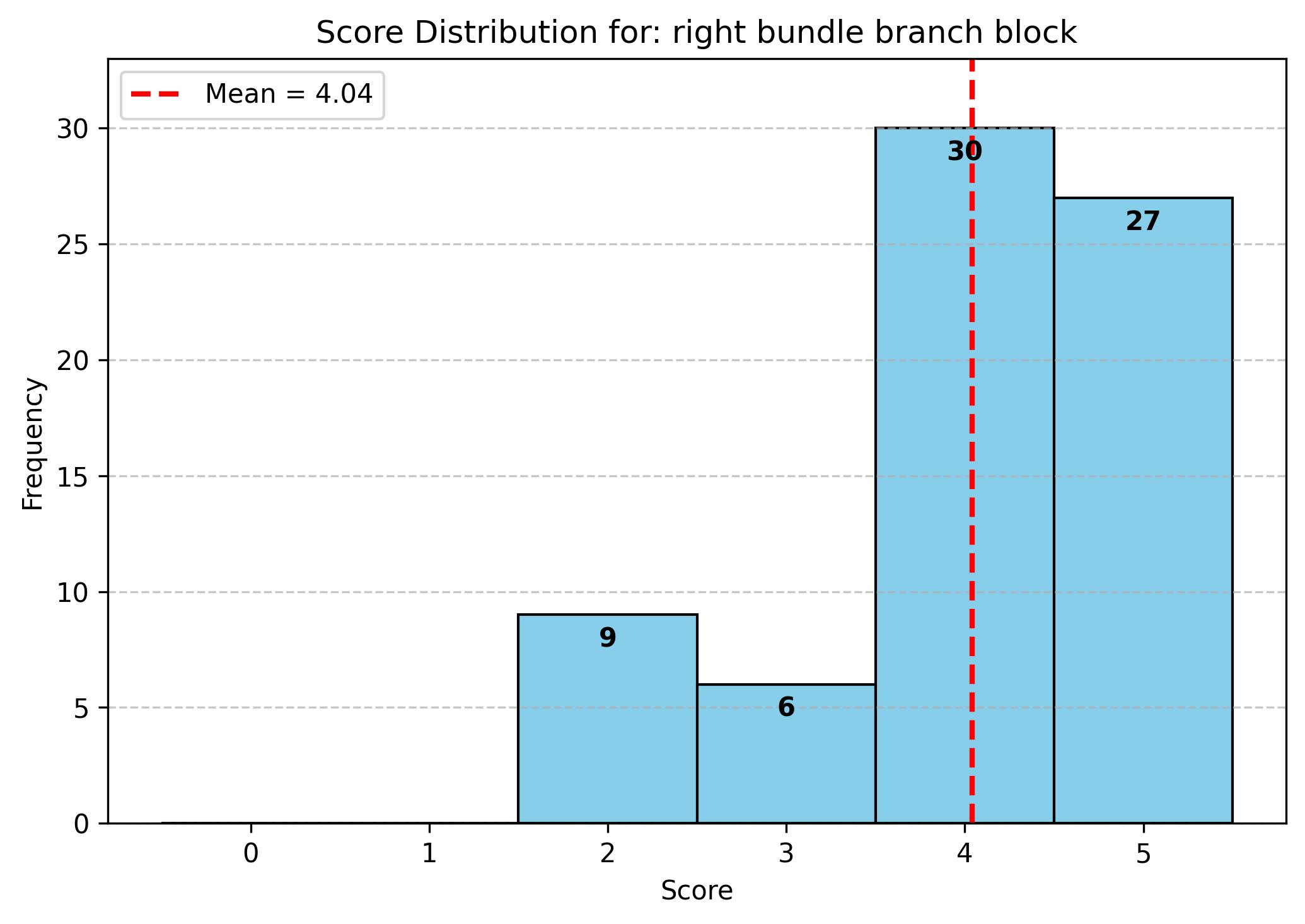}
        \caption{Right bundle branch block}
        \label{fig:rbbb}
    \end{subfigure}

    % Row 3
    \begin{subfigure}{0.48\linewidth}
        \includegraphics[width=\linewidth]{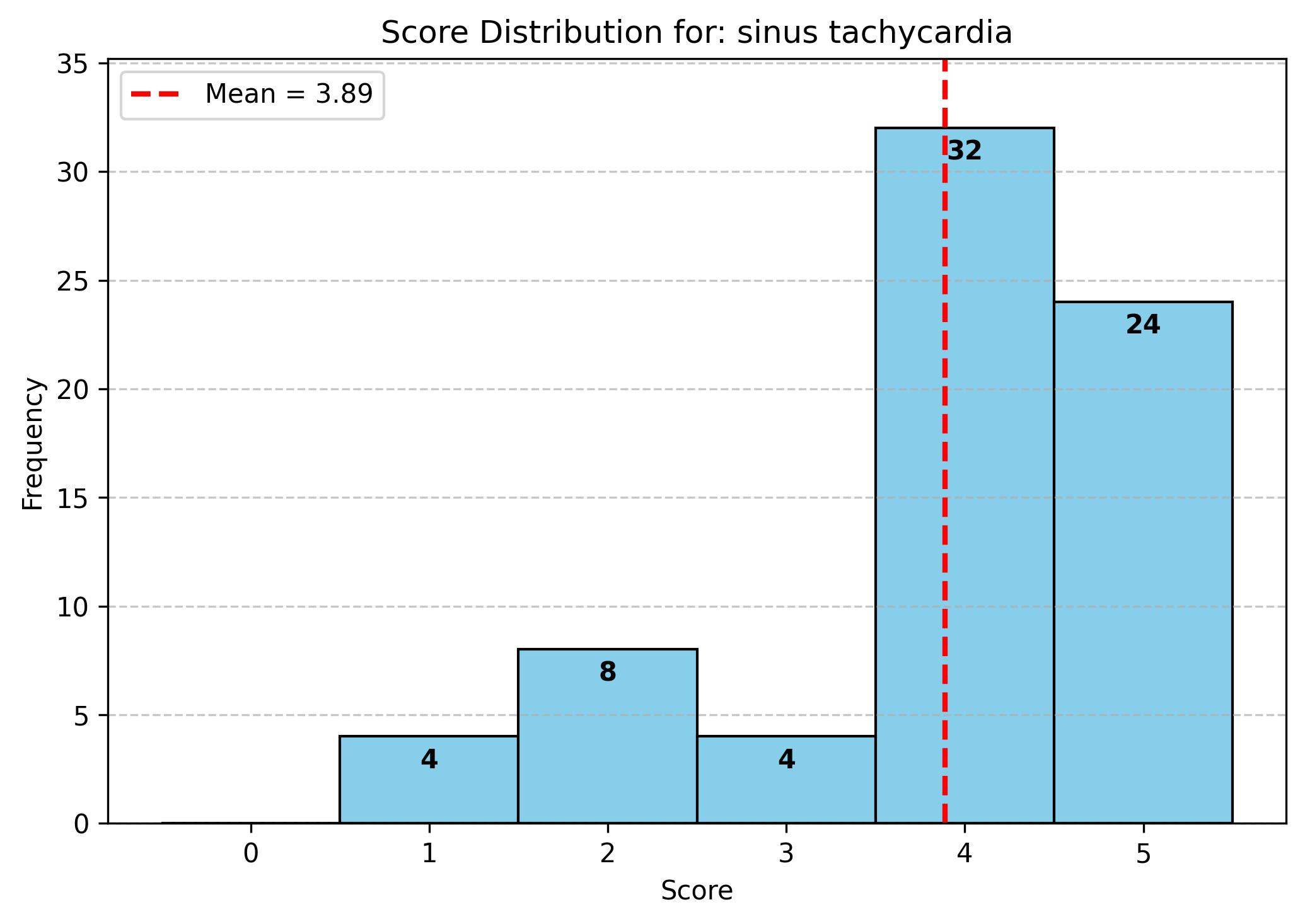}
        \caption{Sinus tachycardia}
        \label{fig:stachy}
    \end{subfigure}
    \hfill
    \begin{subfigure}{0.48\linewidth}
        \includegraphics[width=\linewidth]{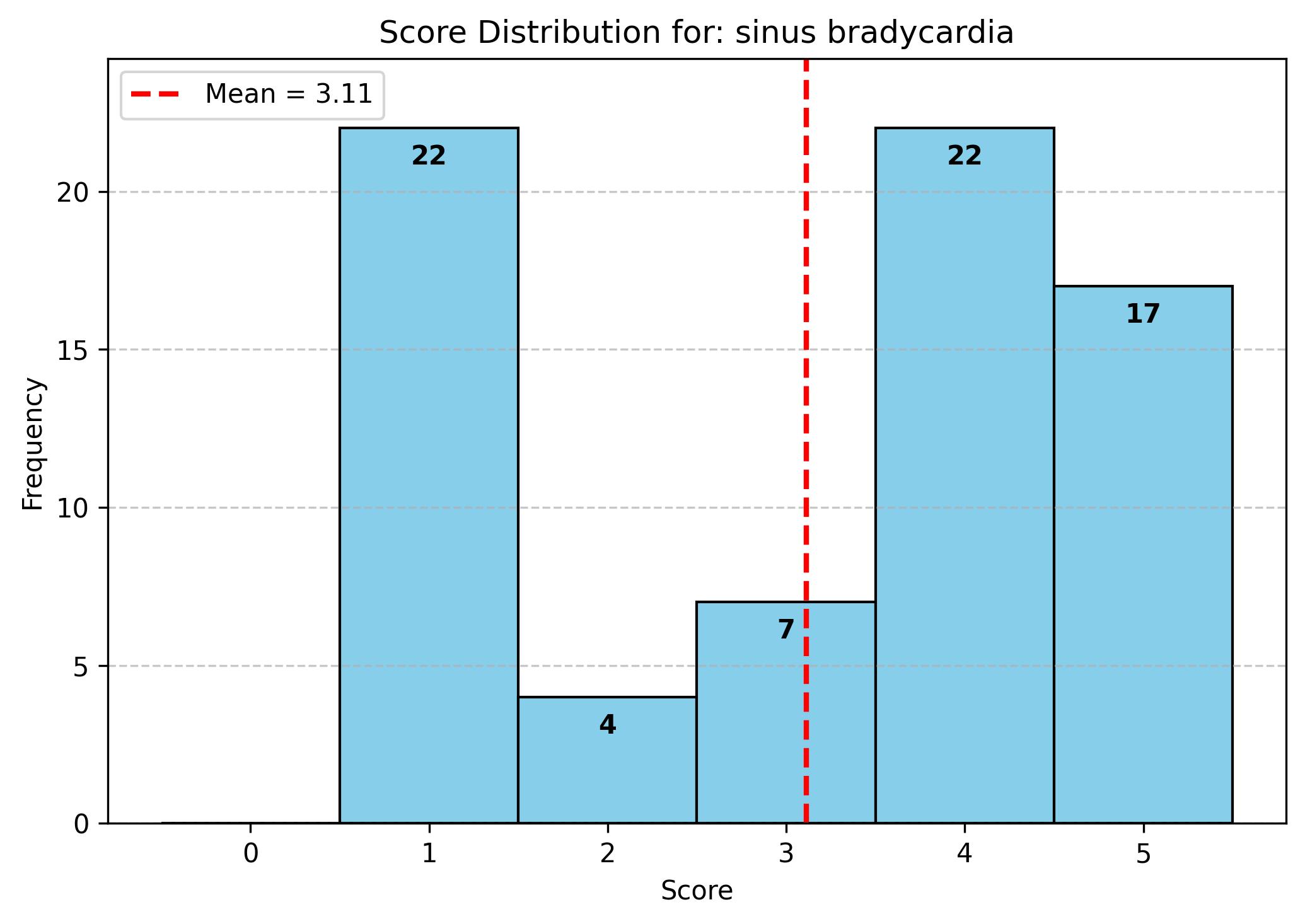}
        \caption{Sinus bradycardia}
        \label{fig:sbrady}
    \end{subfigure}

    \caption{Distribution of scores across expert responses for specific conditions}
    \label{fig:conditions}
\end{figure}

Across all diagnostic conditions, expert scores follow a left-skewed distribution (Fig.~\ref{fig:all}), with a mean of 3.69 and a mode of 4.0 on a five-point scale. These results indicate that in aggregate, experts regard UniCoMTE's explanations as helpful. Figure~\ref{fig:conditions} shows condition-specific score distributions, where the mean rating remains above 3.50 for all except one condition: sinus bradycardia. Explanations for right bundle branch block and sinus tachycardia achieve the highest average scores of 4.04 and 3.89, respectively. This suggests that experts find these counterfactual explanations particularly clear and consistent with expected waveform behavior.
Sinus bradycardia displays a bimodal distribution of ratings (mean 3.11), indicating that some explanations are clear while others provoke disagreement among experts. Among all reviewed cases, only 4 samples receive over half of ratings as “Slightly Misleading” or “Misleading.” Three correspond to sinus bradycardia and one corresponds to left bundle branch block. 

\begin{figure}[H]
\captionsetup{justification=centering}
\centering

% --- Row 1: two subfigures ---
\begin{subfigure}{0.45\linewidth}
    \centering
    \includegraphics[width=\linewidth]{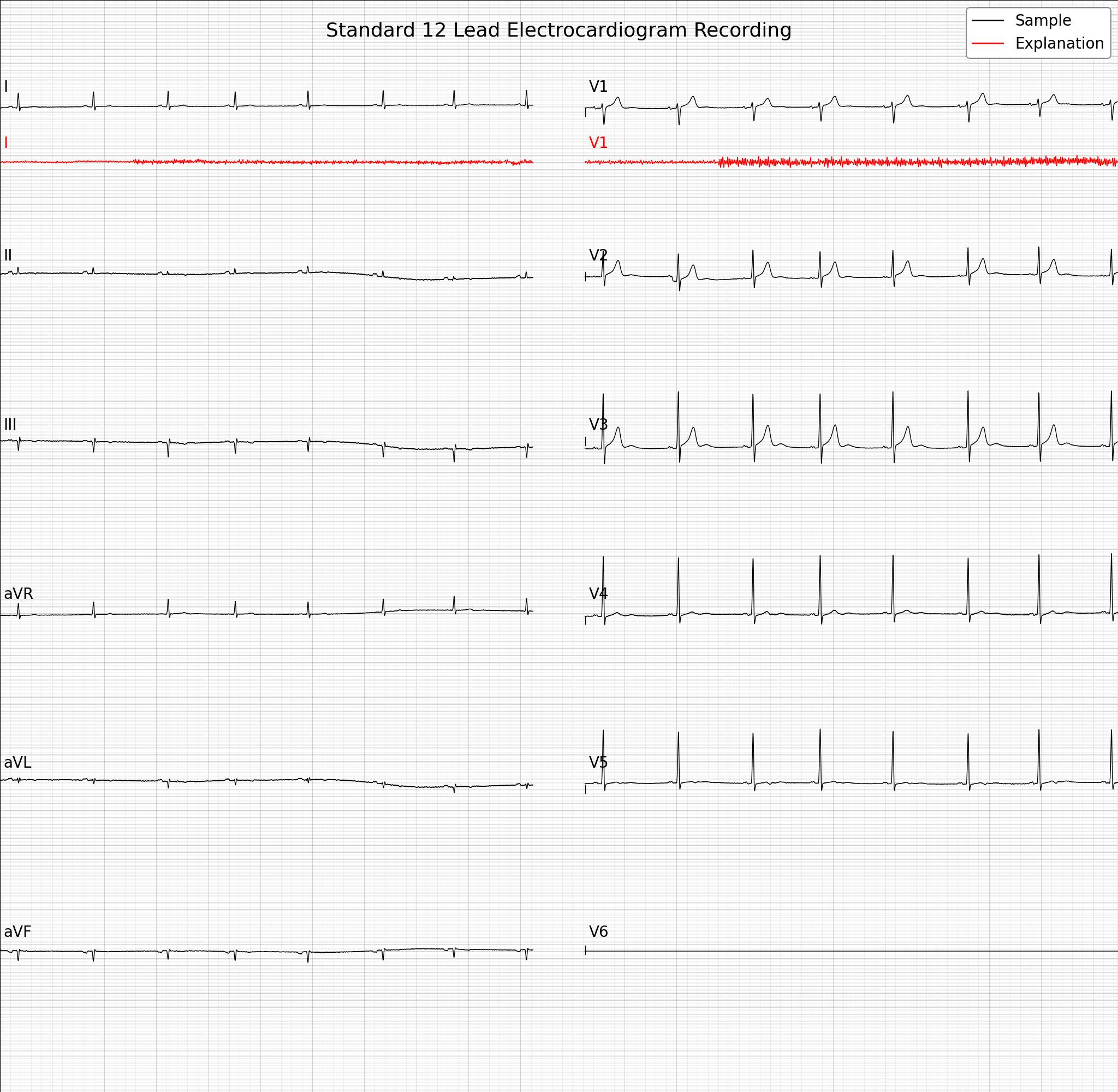}
    \caption{Unhelpful Case 1: Sinus Bradycardia}
    \label{fig:case1}
\end{subfigure}
\hfill
\begin{subfigure}{0.45\linewidth}
    \centering
    \includegraphics[width=\linewidth]{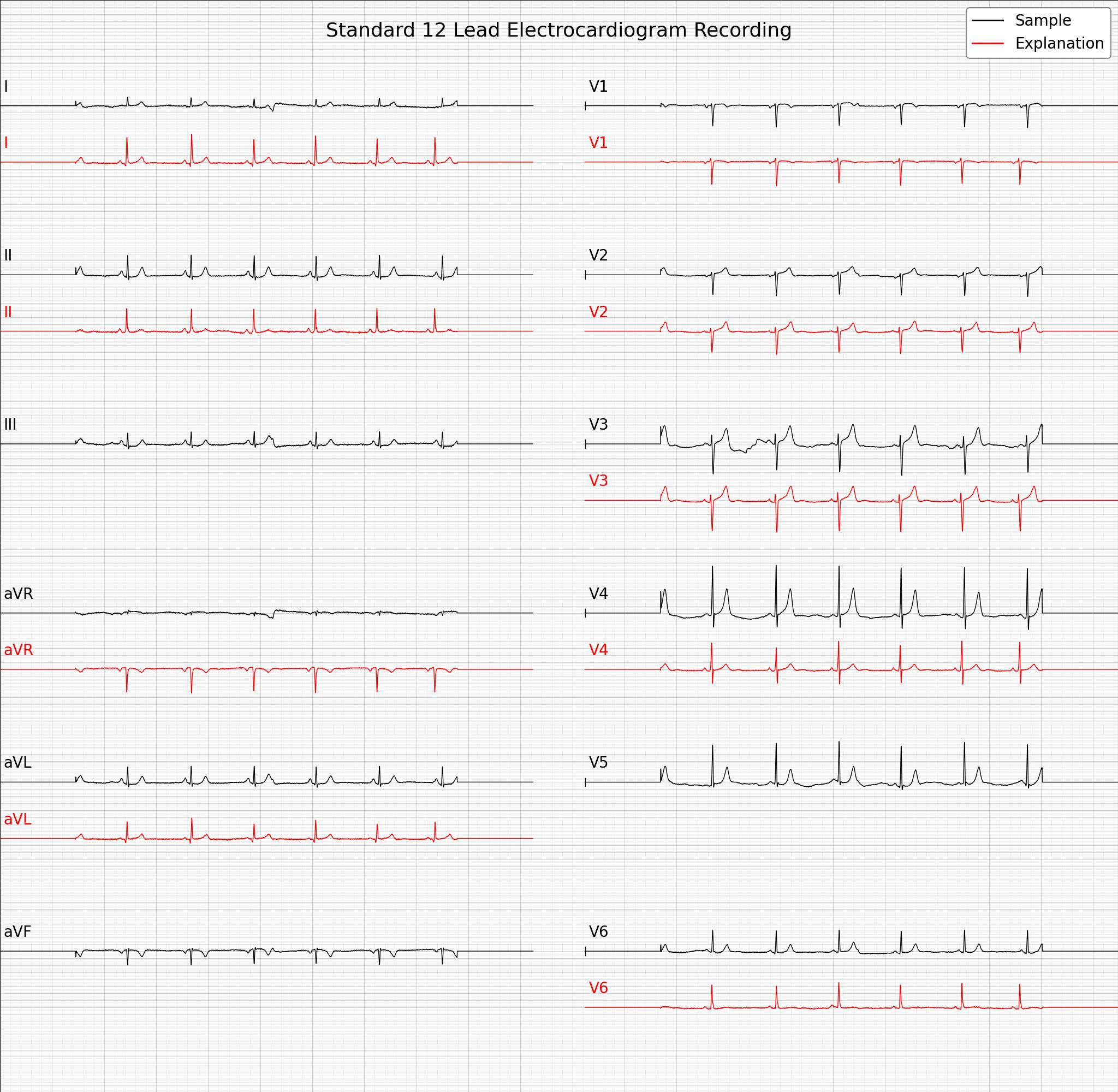}
    \caption{Unhelpful Case 2: Sinus Bradycardia}
    \label{fig:case2}
\end{subfigure}

\vspace{1.5em}

% --- Row 2: two subfigures ---
\begin{subfigure}{0.45\linewidth}
    \centering
    \includegraphics[width=\linewidth]{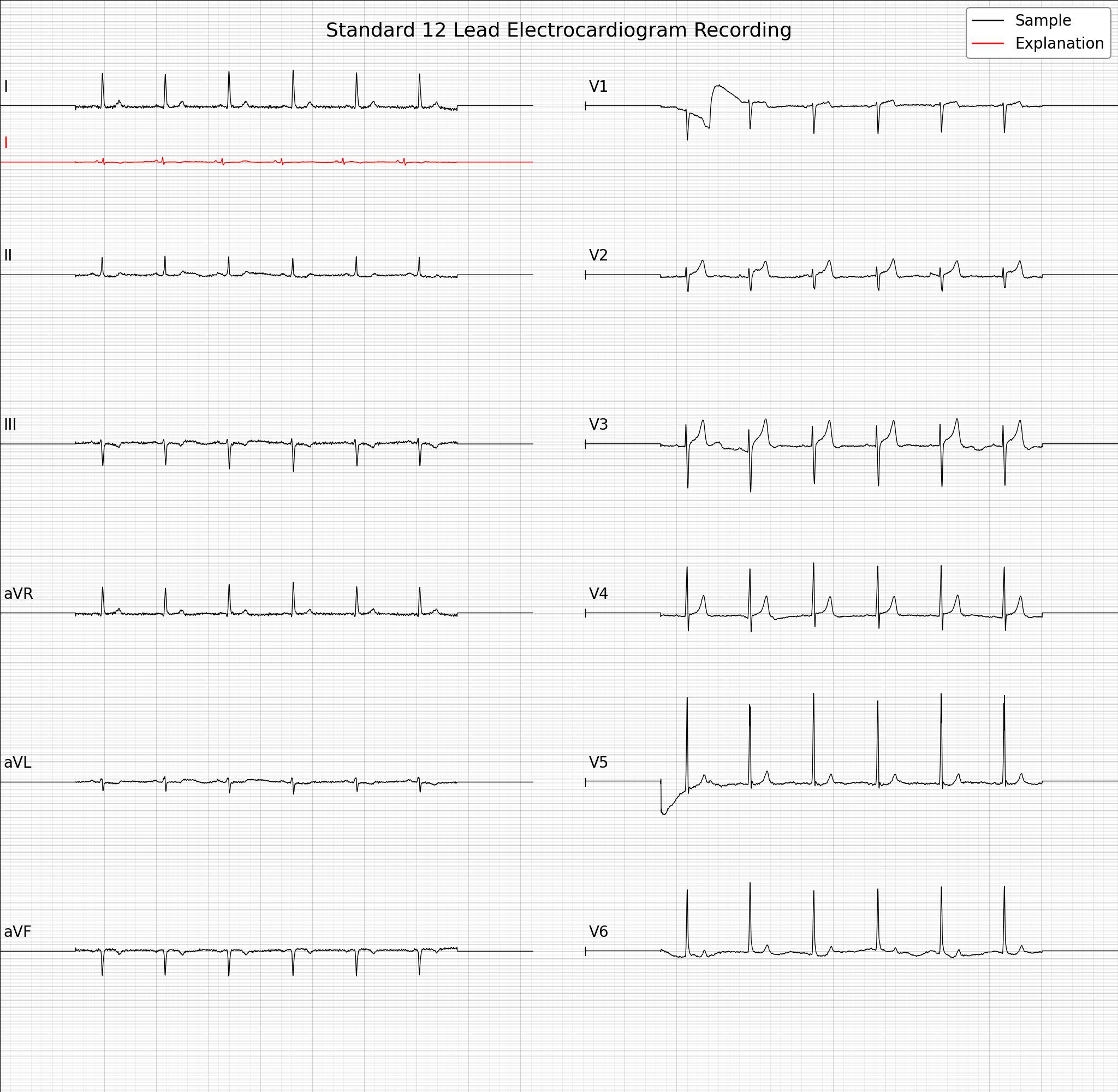}
    \caption{Unhelpful Case 3: Sinus Bradycardia}
    \label{fig:case3}
\end{subfigure}
\hfill
\begin{subfigure}{0.45\linewidth}
    \centering
    \includegraphics[width=\linewidth]{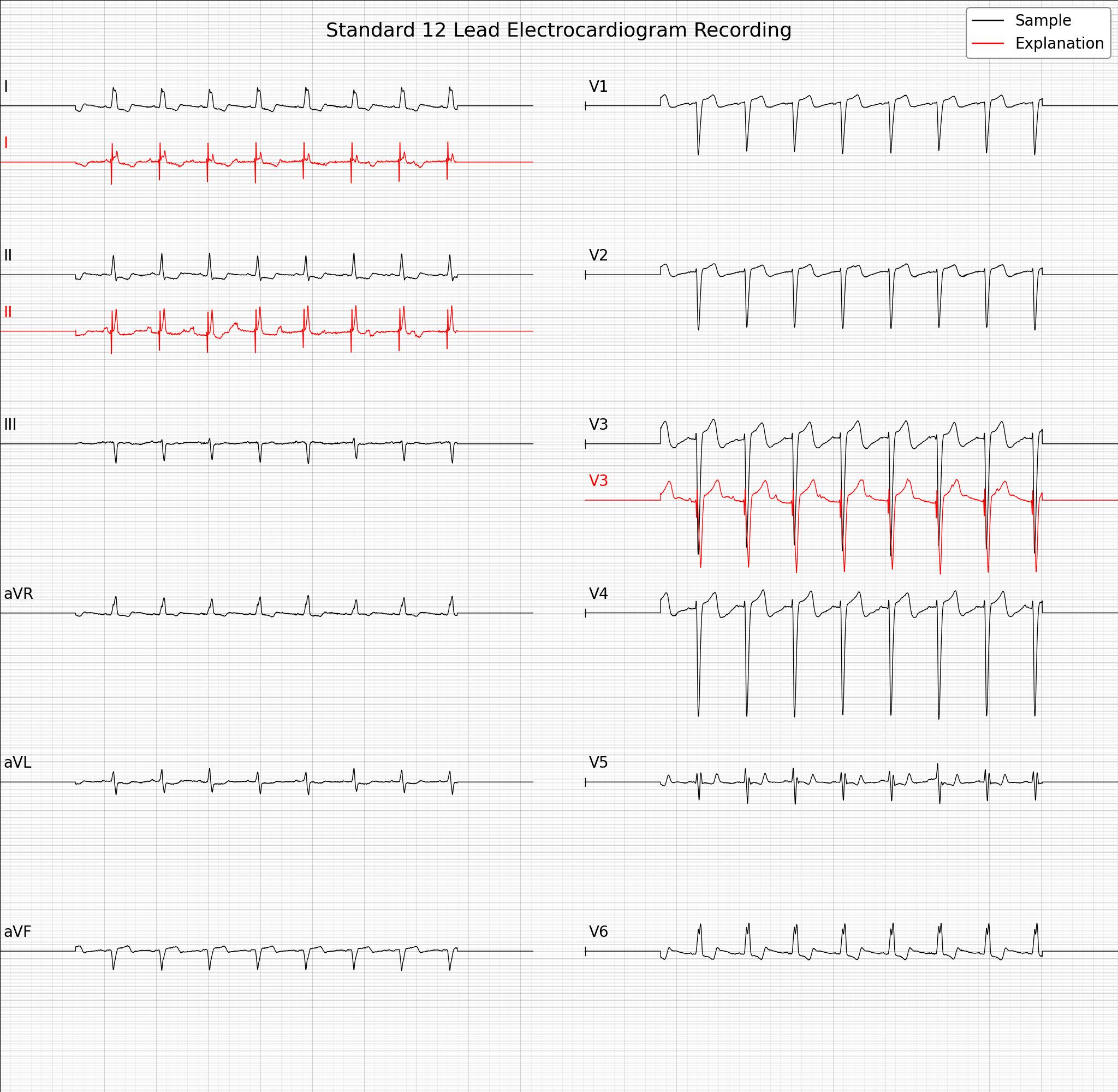}
    \caption{Unhelpful Case 4: Left Bundle Branch Block}
    \label{fig:case4}
\end{subfigure}

\vspace{1.5em}

% --- Row 3: ONE centered subfigure ---
\begin{subfigure}{0.45\linewidth}
    \centering
    \includegraphics[width=\linewidth]{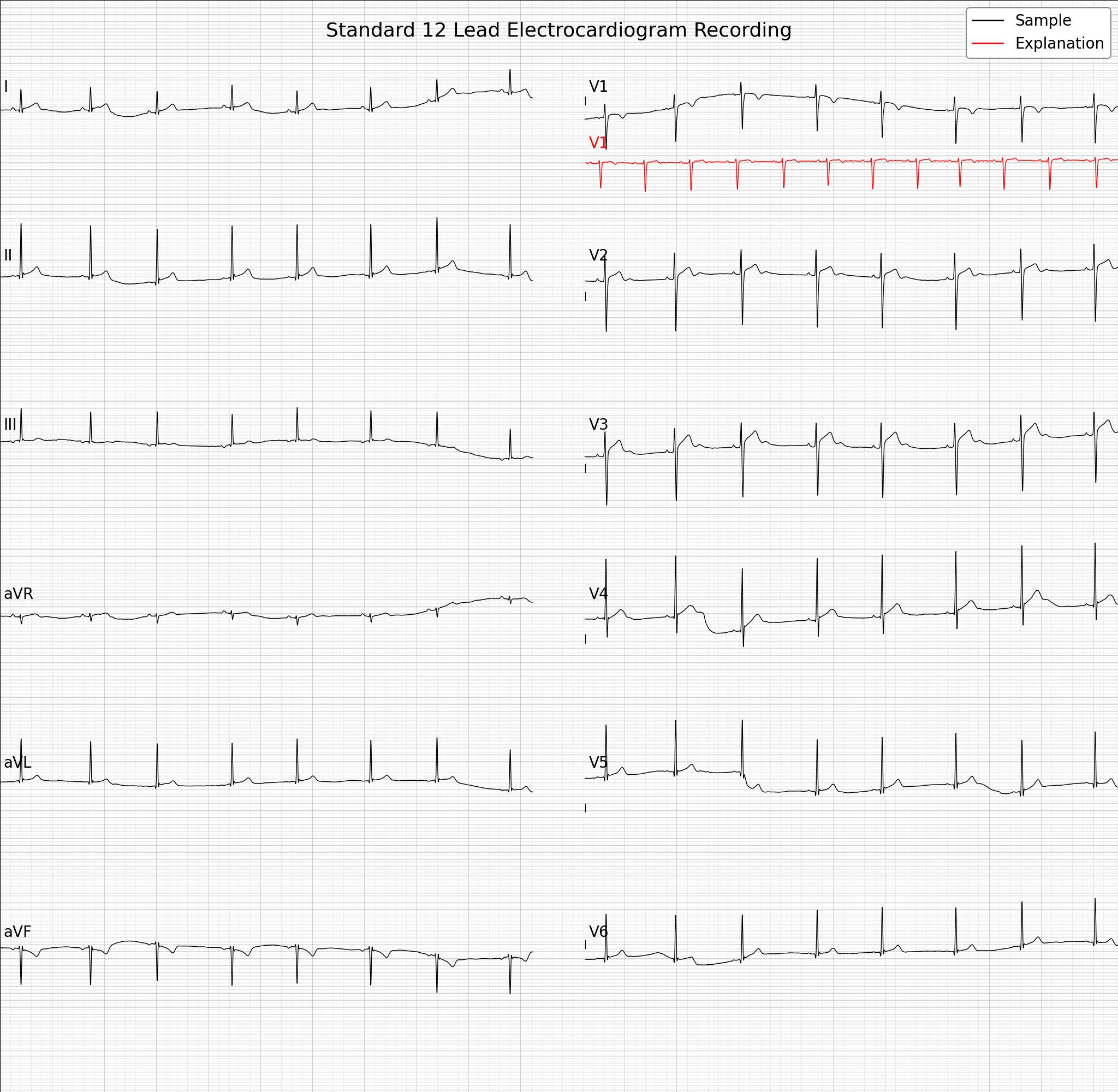}
    \caption{Helpful Case: Sinus Bradycardia}
    \label{fig:case5}
\end{subfigure}

\caption{UniCoMTE counterfactual examples. Original ECGs are shown in black and counterfactuals in red.}
\end{figure}

% Case 1
Figure~\ref{fig:case1} displays the first unhelpful case. The counterfactual includes flatlined or distorted regions that do not contain any clinically meaningful data.  Experts attribute this abnormality to poor signal acquisition or detached leads.
% Case 2 and 3
Figure~\ref{fig:case2} and Figure~\ref{fig:case3}  display the second and third unhelpful explanations that attempt to explain sinus bradycardia. The explanations are misleading because they fail to correct the core abnormality. Both the original signals and explanation signals show a heart rate of approximately 50 bpm, which does not represent a normal heart rate.
% Case 4
Figure~\ref{fig:case4} displays the final unhelpful explanation that attempts to explain LBBB. Experts note that while attempting to correct LBBB, PR intervals are disrupted and atrioventricular block is now present as best seen in lead II.
% Case 5 (helpful) 
In contrast, a highly helpful example (Fig.~\ref{fig:case5}) shows a counterfactual explanation that modifies waveform segments in a physiologically coherent way, producing realistic changes consistent with a healthy rhythm.

Examining the unhelpful samples reveals three primary causes of ineffective explanations. 
First, some counterfactuals appear visually noisy or physiologically implausible due to data-quality issues—such as low-amplitude or flatlined signals—despite prior filtering. Because UniCoMTE draws replacement segments directly from the training set, these failures often reflect poor signal acquisition during dataset construction. 
Second, some explanations fail to correct the original misclassification or inadvertently introduce a different abnormality. This behavior frequently arises from labeling inconsistencies, where an ECG is assigned a label that differs from expert interpretation or where clinicians apply different diagnostic criteria. 
Finally, because UniCoMTE substitutes waveform segments from samples unrelated to the patient being explained, patient-specific morphological differences can yield counterfactuals that appear clinically incoherent or out of context. Some experts also prefer that UniCoMTE include more substituted segments to increase the clarity of the resulting waveform. 
Together, these examples illustrate the current limitations of UniCoMTE’s counterfactual generation and underscore the importance of reliable training data for producing clinically meaningful explanations.

Despite these challenges, experts consistently report that UniCoMTE provides useful visual guidance for understanding how the model distinguishes normal from abnormal ECGs. They note that counterfactual overlays make it easier to identify waveform regions most relevant to the classification, even when the exact physiological correction is imperfect. Experts emphasize that such visual and example-based explanations could serve as effective educational tools for medical trainees or clinicians seeking to understand how machine learning models analyze ECGs. Overall, the qualitative findings support UniCoMTE’s clinical interpretability and highlight its potential to enhance clinicians' trust in time series classifiers used in healthcare. 

\FloatBarrier
\section{Discussion}\label{sec3}

We introduce UniCoMTE, a universal counterfactual explanation framework for multivariate time series, and evaluate its performance on ECG classification. UniCoMTE extends the CoMTE framework with a generalized, model-agnostic architecture that supports diverse ML backends—including TensorFlow, PyTorch, and scikit-learn—and a wide range of model types. This design enables researchers to apply a single counterfactual explainability workflow across architectures and domains without model-specific adjustments. The framework emphasizes flexibility, interpretability, and practical integration into existing time-series analysis pipelines.

UniCoMTE advances interpretability by producing concise and actionable counterfactuals that specify the minimal signal segments whose modification would change a model’s decision. In contrast to LIME and SHAP, which yield dense data point level coefficients, UniCoMTE generates localized, feature-level signal adjustments that more naturally align with human reasoning. This approach allows users to visualize what alterations in waveform patterns lead to a change in classification, offering an intuitive, example-driven view of model behavior. The quantitative analyses highlight two key properties—\textit{comprehensibility} and \textit{generalizability}. UniCoMTE consistently identifies a small number of relevant features, making its explanations easier to interpret than attribution-based alternatives. Furthermore, counterfactuals generalize across multiple misclassified samples of the same type, capturing systematic trends influencing model decision-making, rather than isolated errors. This property positions UniCoMTE as both a local explanation method and a tool for global model auditing.

The expert evaluation underscores the practical value of counterfactual reasoning for clinical interpretability. Clinicians report that UniCoMTE’s explanations resemble the “what-if” reasoning process that underpins diagnostic decision-making. By visualizing minimally altered ECGs that revert an abnormal prediction to normal, the framework bridges the gap between abstract model outputs and clinically meaningful evidence. As a result, this framework has the potential to help build clinician trust and understanding in model interpretation.
Beyond model interpretability, UniCoMTE aids dataset validation and quality assurance. The framework exposes samples with missing or corrupted physiological signals and highlights inconsistencies in training labels, providing feedback that can improve data curation and model reliability. These diagnostic capabilities suggest broader utility in refining datasets used for clinical machine learning.

Despite these advantages, several challenges remain. Implausible counterfactuals occasionally arise from noisy or mislabeled training samples, underscoring the need for rigorous data verification. Expert disagreement across some conditions further reflects the intrinsic subjectivity of cardiac diagnosis. In addition, the current hill-climbing search procedure imposes computational overhead, which may constrain large-scale or real-time deployment.
Future work can extend UniCoMTE by integrating physiologically informed constraints to ensure signal plausibility and by incorporating uncertainty estimates to quantify the reliability of generated counterfactuals. Active learning~\cite{budd2021survey, konyushkova2017learning, biswas2023active} strategies could further enhance dataset robustness by iteratively identifying and correcting mislabeled samples. Owing to its model-agnostic and domain-independent design, UniCoMTE can also generalize beyond ECGs to other time-series domains such as EEG~\cite{das2023survey}, wearable monitoring~\cite{takei2015toward}, or industrial IoT telemetry~\cite{de2020novel}. By unifying architectural flexibility with clinically validated interpretability, UniCoMTE provides a blueprint for trustworthy and broadly applicable counterfactual reasoning in time-series analysis.

\section{Methods}\label{sec4}
\subsection{ECG Dataset and Preprocessing}
We use the publicly available CODE-15~\cite{ribeiro_2021_4916206} and CODE-test~\cite{ribeiro_2020_3765780} datasets, released as part of prior work on automated ECG diagnosis. CODE-15 serves as the training set and contains 345{,}779 12-lead ECG recordings, while CODE-test, used for evaluation, comprises 827 recordings. All signals are sampled at 400~Hz, and we apply zero-padding to standardize each lead to 4{,}096 samples (approximately 10~seconds of recording).
Each ECG is annotated with one or more of six diagnostic classes: first-degree atrioventricular block (1dAVb), right bundle branch block (RBBB), left bundle branch block (LBBB), sinus bradycardia (SB), atrial fibrillation (AF), and sinus tachycardia (ST). Labels are provided in one-hot encoded form.

Annotation procedures differ across the two datasets. In CODE-test, cardiologists assign labels by consensus, with a senior expert resolving disagreements. In CODE-15, diagnoses are assigned by a semi-automated pipeline that integrates structured outputs from the University of Glasgow ECG analysis system (Uni-G), automatic signal measurements, and text mining of cardiologist reports.
We adopt these datasets because they support the state-of-the-art CNN model evaluated in this study, and we use them directly without additional normalization, resampling, or feature extraction.

\begin{figure}[t!]
    \centering
    \includegraphics[width=0.85\linewidth]{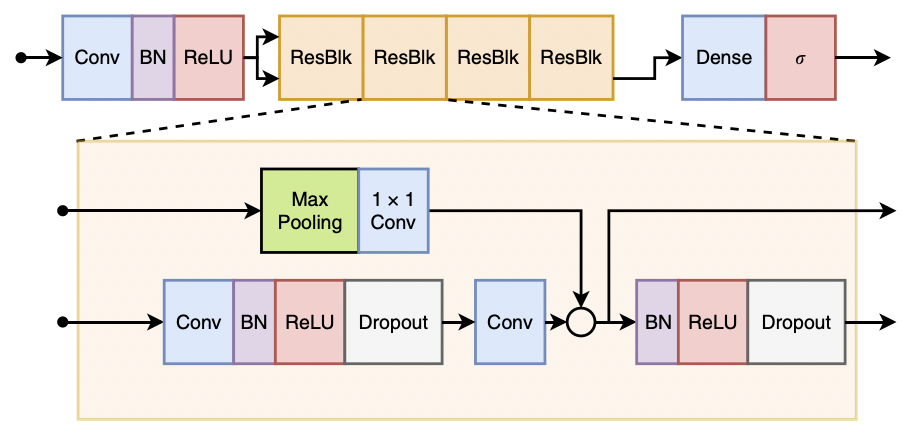}
    \caption{The deep neural network architecture for ECG classification adopted from Ribeiro et al.~\cite{ribeiro2020automatic}. The model applies convolutional and residual layers to extract temporal patterns from 12-lead ECG signals.}
    \label{fig:cnn_model}
\end{figure}

\subsection{Classification Model}
We adopt the CNN architecture proposed by Ribeiro et al.~\cite{ribeiro2020automatic} for ECG classification. The model takes raw 12-lead ECG time series as input and predicts six diagnostic categories (1dAVb, RBBB, LBBB, SB, AF, and ST). Each input sample contains 12 leads, zero-padded to 4{,}096 samples (approximately 10~seconds at 400~Hz). The network follows a one-dimensional residual architecture adapted from ResNet~\cite{targ2016resnet}: an initial convolutional layer is followed by four residual blocks, each containing two convolutional layers with batch normalization and ReLU activation. Max-pooling layers reduce temporal resolution between blocks, while skip connections preserve information flow. A fully connected layer with sigmoid activation produces the six-dimensional probability output, and dropout after ReLU layers improves generalization. The overall model architecture is shown in Figure~\ref{fig:cnn_model}.

We use the pretrained instance released by Ribeiro et al., selecting the model that achieves a micro-average precision of 0.951 across ten independent runs. After generating predictions, we apply the class-specific thresholds defined in their work, which maximize the F1-score on the validation set and yield positive class predictions for each condition.
The resulting CNN achieves strong baseline performance on the CODE-test dataset, which comprises 827 expert-annotated ECGs. Precision, recall, and F1-scores remain high across most diagnostic categories, with a modest reduction in recall for AF. Table~\ref{tab:cnn_performance} summarizes the class-wise results. This CNN serves as the target classifier for generating and evaluating UniCoMTE explanations throughout the study.

\begin{table}[t!]
\centering
\caption{Performance of the CNN model on the CODE-test set.} 

\label{tab:cnn_performance}
\begin{tabular}{lccc}
\toprule
\textbf{Condition} & \textbf{Precision} & \textbf{Recall} & \textbf{F1 Score} \\
\midrule
Normal & 0.99 & 0.98 & 0.99 \\
1dAVb & 1.00 & 0.64 & 0.78 \\
RBBB & 0.87 & 0.97 & 0.92 \\
LBBB & 0.97 & 1.00 & 0.98 \\
SB & 0.78 & 0.88 & 0.82 \\
AF & 1.00 & 0.54 & 0.70 \\
ST & 0.92 & 0.92 & 0.92 \\
\bottomrule
\end{tabular}
\end{table}

% This model achieves F1 scores above 80\% and specificity over 99\% across all six diagnostic classes on the CODE-test dataset. Given its strong performance and relevance to clinical ECG interpretation, we use this model as the target for generating CoMTE-V1.1 explanations.

% CNN architecture (depth, layers, activation functions)
% Training parameters (loss function, optimizer, batch size, epochs)
% Justification for why this is a strong baseline

\subsection{UniCoMTE Framework}
We develop UniCoMTE, a model-agnostic framework for generating counterfactual explanations in multivariate time series classifiers. Similar to the original CoMTE method~\cite{ates2021counterfactual}, UniCoMTE identifies the temporal and variable segments that most influence a model’s prediction. This is done by constructing a minimally modified sample that is classified as a target class, rather than the original class. Unlike its predecessor, we design UniCoMTE as a modular and extensible pipeline that supports diverse ML backends—including scikit-learn, PyTorch, and TensorFlow—without model-specific adaptation.

The framework comprises three main components: a \textit{data and model wrapper}, a \textit{distractor selection module}, and a \textit{counterfactual generation module}. Given a trained classifier and a sample of interest, UniCoMTE first retrieves \textit{distractors}—samples from the target class that are as similar as possible to the input. This search employs class-specific KD-trees~\cite{procopiuc2003bkd} constructed from correctly classified training examples, enabling efficient nearest-neighbor retrieval.

The wrapper subsystem abstracts out differences in model architectures and data representations. The \textit{model wrapper} provides a unified interface that allows the user to standardize how UniCoMTE queries model predictions and class probabilities, regardless of the underlying backend and complexity of the classification algorithm. Wrappers enable the user to define the function call that suits their classifier and incorporate relevant pre/post processing operations such as thresholding. This abstraction enables UniCoMTE to access and manipulate any black-box classifier, avoiding the need for manual reconfiguration. 

The \textit{data wrapper} allows the user to define how their raw time series arrays can be reformatted into \textit{pandas} MultiIndex DataFrames indexed by both entity (e.g., ECG lead, sensor, or compute node) and time, with columns corresponding to measured variables for each entity. This structure enables efficient slicing of multivariate sequences and selective replacement of specific variable–time segments during counterfactual construction. 

Using the wrapped data and model, UniCoMTE performs a discrete random hill-climbing search~\cite{jacobson2004analyzing} to identify the smallest set of variable–time pairs that, when replaced with corresponding values from a distractor, change the model’s output to the target class with maximal confidence. If this optimization fails to find a valid counterfactual, the framework reverts to a greedy incremental strategy that tests single-feature replacements and expands the candidate set iteratively.

\begin{figure}[htbp]
    \centering
    \includegraphics[width=0.9\linewidth]{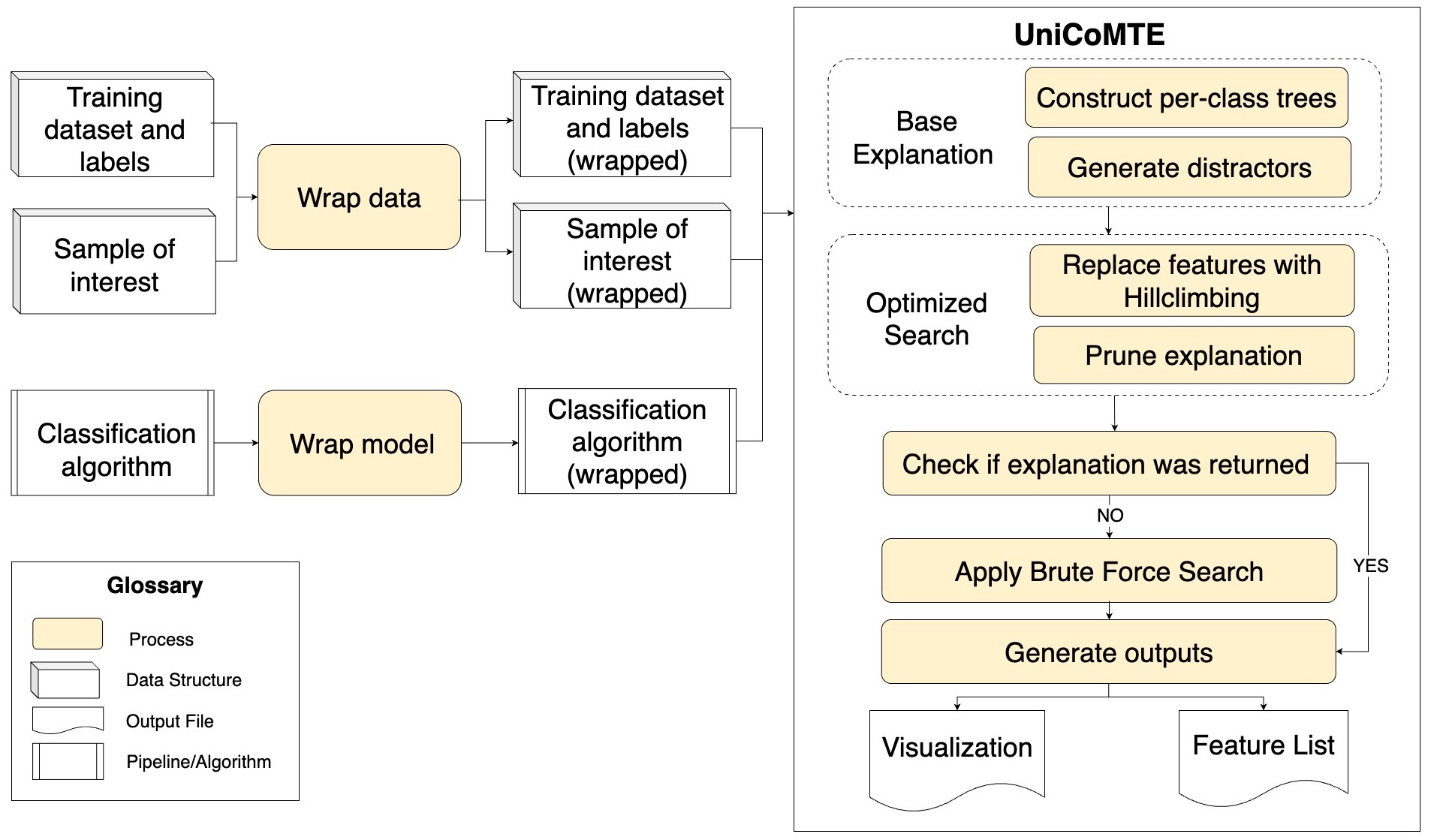}
    \caption{High level architecture of UniCoMTE.}
    \label{fig:comteV1.1-architecture}
\end{figure}

The resulting counterfactual specifies the minimal set of time–lead segments and their replacement values required to alter the model’s decision. These explanations are both sparse and actionable, revealing not only \textit{which} features are critical but also \textit{how} they must change to influence classification. In the context of ECG analysis, UniCoMTE allows clinicians to visualize waveform regions that drive predictions, bridging the gap between machine inference and physiologically interpretable reasoning. Figure~\ref{fig:comteV1.1-architecture} summarizes the overall framework.

\subsection{LIME and SHAP Comprehensibility Implementation}

To evaluate comprehensibility, we apply UniCoMTE on all CODE-test samples that are classified as abnormal (149 samples), and set UniCoMTE's class of interest to the normal class. We record the number of features returned for each sample, and record the mean and mode across the 149 samples. 

We also apply LIME and SHAP on a randomly selected subset of the test dataset for comparison with UniCoMTE. We implement custom data and model wrappers to ensure compatibility with both libraries. For SHAP, we use the \texttt{GradientExplainer} module, and for LIME, the \texttt{TabularExplainer} module, as the ECG data are not suitable for the image- or text-based variants. We use the default configuration of each library, allowing LIME to return the ten most influential features in each explanation. For each method, we record the number of unique features reported per explanation, which represents the amount of information a user must process to interpret the model’s decision.

\subsection{Assessment of Generalizability}
We evaluate generalizability by testing whether a counterfactual generated for one misclassified sample can also correct other misclassifications of the same type. For each error category (e.g., Normal predicted as 1dAVb), we generate an explanation from a representative sample and apply the same feature substitutions to all remaining misclassified samples within that category. Each case in which the prediction flips to the correct label is counted as a successful correction. We quantify generalizability using \textit{coverage}, defined as the proportion of successful corrections among all tested misclassifications of that type.

To ensure a sufficient number of test cases, we augment the evaluation set with the first 3{,}000 samples from the training dataset, removing them beforehand to avoid data leakage. We focus the analysis on the six Normal-class misclassification categories with the highest sample counts, as understanding why the model incorrectly predicts a normal tracing as an abnormal one is a clinically relevant scenario.

\subsection{Design of Expert Questionnaire}

We design an expert questionnaire to assess the clarity and clinical utility of counterfactual explanations produced by UniCoMTE. For this evaluation, we select the first 47 correctly classified abnormal samples from the test dataset—eight for each of the six diagnostic conditions, except atrial fibrillation, which contains seven positive samples. In all cases, the Normal class serves as the counterfactual reference, representing the clinically meaningful scenario in which a physician seeks to understand why the model predicts an abnormal tracing instead of a normal one.

To visualize the ECG data, we adapt an open-source 12-lead ECG plotting tool and extend it to overlay counterfactual explanations in red beneath the original traces. This design allows experts to directly compare the observed and counterfactual signals and assess how specific waveform modifications drive the model’s prediction toward the Normal class. Alongside each plot, we display the model’s original prediction and the counterfactual target.

During explanation generation, we observe that some distractor samples exhibit flatlined or non-physiological signals, likely caused by sensor faults or data collection errors. To mitigate this issue, we apply a quality-control filter that removes Normal-class samples with near-zero variability. We flatten each signal into a vector and exclude samples whose standard deviation falls below 0.1. This procedure eliminates 821 low-quality signals from the training dataset, improving the reliability of generated counterfactuals.

The final questionnaire contains 60 items, each corresponding to one diseased sample. For each item, a clinical expert reviews the ECG plot with the overlaid counterfactual, considers the model’s prediction, and rates how well the explanation clarifies the model’s reasoning. We collect responses on a five-point Likert scale, where higher scores indicate greater perceived clarity and interpretability. We recruit seven practicing clinicians specializing in cardiology to complete the evaluation. Each participant provides informed consent before participation. Their feedback directly measures whether UniCoMTE produces explanations that align with expert diagnostic reasoning and clinical expectations.

\subsection{Code availability}
The UniCoMTE implementation, along with model and data wrappers and experiment scripts, is publicly available at:
https://github.com/peaclab/UniCoMTE/tree/main.

\subsection{Competing interests}
The authors declare no competing interests.

% \subsection{Ethics declarations}
% This study uses de-identified, publicly available ECG datasets (CODE-15 and CODE-test) and does not involve interventions or access to protected health information. The expert questionnaire targets clinicians reviewing synthetic visualizations and does not collect sensitive personal data. Institutional review determined the survey to be exempt/non-human-subjects research. 

\subsection{Acknowledgements}
We gratefully acknowledge the clinicians who participated in the expert questionnaire and provided invaluable feedback on the UniCoMTE explanations: Dr.~Stephen Tsaur (Boston Medical Center, MA), Dr.~Tae Kyung Yoo (Boston Medical Center, Department of Cardiology, MA), Dr.~Murat M.~Yilmazer (Behçet Uz Children’s, Turkey), Dr.~Nurseli Bayram and Dr.~Emir Ünal (Marmara University School of Medicine, Turkey), Dr.~Nirupama Vellanki Mithal (Boston Medical Center, Department of Cardiology, MA), Dr.~Lucas Casul (Presbyterian Medical Group, NM), Dr.~Caroline Kaufman (Boston Medical Center, MA), and Dr.~Jamel Ortoleva (Boston Medical Center, Department of Anesthesiology, MA). Their insights were essential for assessing the clinical relevance and interpretability of the generated explanations. 
We also thank Dr.~Emre Ates and Dr.~Burak Aksar for developing the original CoMTE framework, which laid the groundwork for the UniCoMTE methodology presented in this work.

This work has been partially funded by Sandia National Laboratories. The work to support TensorFlow in UniCoMTE was funded by the U.S. Department of Energy National Nuclear Security Administration’s Office of
Defense Nuclear Nonproliferation Research and Development. Sandia National Laboratories is a multimission laboratory managed and operated by National Technology and Engineering Solutions of Sandia, LLC., a wholly owned subsidiary of Honeywell International, Inc., for the U.S. Department of Energy’s National Nuclear Security Administration under Contract DE-NA0003525. This paper describes objective technical results and analysis. Any subjective views or opinions that might be expressed in the paper do not necessarily represent the views of the U.S. Department of Energy or the United States Government.

\subsection{Author contributions}
A.K.C. and V.J.L. conceived and supervised the study. J.L. implemented the software, and ran experiments. J.Z.D. contributed to method design, engineering of the wrappers of the UniCoMTE framework. J.L. and E.S. prepared the figures, generated the quantitative and qualitative analyses. J.L. and E.S. coordinated the expert questionnaire and curated the explanation visualizations. J.L. and E.S. wrote the initial manuscript draft; J.Z.D., V.J.L., S.T, and A.K.C. reviewed, edited, and refined the manuscript. V.J.L. and A.K.C. provided project guidance and resources. All authors discussed the results, interpreted findings, and approved the final manuscript.

% Number and background of clinicians
\bibliography{sn-bibliography}% common bib file

@inproceedings{ates2021counterfactual,
  title={Counterfactual explanations for multivariate time series},
  author={Ates, Emre and Aksar, Burak and Leung, Vitus J and Coskun, Ayse K},
  booktitle={2021 international conference on applied artificial intelligence (ICAPAI)},
  pages={1--8},
  year={2021},
  organization={IEEE}
}

@article{jang2025gcx,
  author    = {Jang, J. H. and Jo, Y. Y. and Kang, S. and others},
  title     = {A novel XAI framework for explainable AI-ECG using generative counterfactual XAI (GCX)},
  journal   = {Scientific Reports},
  volume    = {15},
  pages     = {23608},
  year      = {2025},
  doi       = {10.1038/s41598-025-08080-5},
  url       = {https://doi.org/10.1038/s41598-025-08080-5}
}

@article{mertes2022ganterfactual,
  author    = {Mertes, Silvan and Huber, Tobias and Weitz, Katharina and Heimerl, Alexander and Andr{\'e}, Elisabeth},
  title     = {GANterfactual---Counterfactual Explanations for Medical Non-experts Using Generative Adversarial Learning},
  journal   = {Frontiers in Artificial Intelligence},
  volume    = {5},
  pages     = {825565},
  year      = {2022},
  doi       = {10.3389/frai.2022.825565},
  url       = {https://doi.org/10.3389/frai.2022.825565}
}

@article{SINGLA2023102721,
title = {Explaining the black-box smoothly—A counterfactual approach},
journal = {Medical Image Analysis},
volume = {84},
pages = {102721},
year = {2023},
issn = {1361-8415},
doi = {https://doi.org/10.1016/j.media.2022.102721},
url = {https://www.sciencedirect.com/science/article/pii/S1361841522003498},
author = {Sumedha Singla and Motahhare Eslami and Brian Pollack and Stephen Wallace and Kayhan Batmanghelich}
}

@misc{who-cvd,
  author       = {{World Health Organization}},
  title        = {Cardiovascular diseases (CVDs)},
  year         = {2021},
  url          = {https://www.who.int/news-room/fact-sheets/detail/cardiovascular-diseases-(cvds)},
  note         = {Accessed: 2025-04-15}
}

@article{hannun2019cardiologist,
  title={Cardiologist-level arrhythmia detection and classification in ambulatory electrocardiograms using a deep neural network},
  author={Hannun, Awni Y and Rajpurkar, Pranav and Haghpanahi, Mohammad and Tison, Geoffrey H and Bourn, Carly and Turakhia, Mintu P and Ng, Andrew Y},
  journal={Nature Medicine},
  volume={25},
  number={1},
  pages={65--69},
  year={2019},
  publisher={Nature Publishing Group}
}

@article{attia2019artificial,
  title={An artificial intelligence-enabled ECG algorithm for the identification of patients with atrial fibrillation during sinus rhythm: a retrospective analysis of outcome prediction},
  author={Attia, Zachi I and Friedman, Paul A and Noseworthy, Peter A and Lopez-Jimenez, Francisco and Ladewig, Daniel J and Satam, Gautham and Pellikka, Patricia A and Munger, Thomas M and Asirvatham, Samuel J and Scott, Christopher G and others},
  journal={The Lancet},
  volume={394},
  number={10201},
  pages={861--867},
  year={2019},
  publisher={Elsevier}
}

@article{rajpurkar2017cardiologist,
  title={Cardiologist-level arrhythmia detection with convolutional neural networks},
  author={Rajpurkar, Pranav and Hannun, Awni and Haghpanahi, Mohammad and Bourn, Carly and Ng, Andrew Y},
  journal={arXiv preprint arXiv:1707.01836},
  year={2017}
}

@article{ribeiro2020automatic,
  title={Automatic diagnosis of the 12-lead ECG using a deep neural network},
  author={Ribeiro, Ant{\^o}nio H and Ribeiro, Manoel Horta and Paix{\~a}o, Gabriela MM and Oliveira, Derick M and Gomes, Paulo R and Canazart, J{\'e}ssica A and Ferreira, Milton PS and Andersson, Carl R and Macfarlane, Peter W and Meira Jr, Wagner and others},
  journal={Nature communications},
  volume={11},
  number={1},
  pages={1760},
  year={2020},
  publisher={Nature Publishing Group UK London}
}

@article{o2015introduction,
  title={An introduction to convolutional neural networks},
  author={O'shea, Keiron and Nash, Ryan},
  journal={arXiv preprint arXiv:1511.08458},
  year={2015}
}

@article{shiri2023comprehensive,
  title={A comprehensive overview and comparative analysis on deep learning models: CNN, RNN, LSTM, GRU},
  author={Shiri, Farhad Mortezapour and Perumal, Thinagaran and Mustapha, Norwati and Mohamed, Raihani},
  journal={arXiv preprint arXiv:2305.17473},
  year={2023}
}

@article{alamatsaz2022lightweight,
  title={A lightweight hybrid cnn-lstm model for ecg-based arrhythmia detection},
  author={Alamatsaz, Negin and Yazdchi, Mohammadreza and Payan, Hamidreza and Alamatsaz, Nima and Nasimi, Fahimeh and others},
  journal={arXiv preprint arXiv:2209.00988},
  year={2022}
}

@article{aziz2021ecg,
  title={ECG-based machine-learning algorithms for heartbeat classification},
  author={Aziz, Saira and Ahmed, Sajid and Alouini, Mohamed-Slim},
  journal={Scientific reports},
  volume={11},
  number={1},
  pages={18738},
  year={2021},
  publisher={Nature Publishing Group UK London}
}

@article{perturbation2022,
  title={Perturbation-based explainable AI for ECG sensor data},
  author={Liu, Qi and Yan, Wen and Wang, Zhenyu and Li, Yueming},
  journal={Biomedical Signal Processing and Control},
  volume={75},
  pages={103584},
  year={2022},
  publisher={Elsevier}
}

@article{shap_hrv,
  title={Explainable artificial intelligence for heart rate variability in ECG signal},
  author={Acharya, U Rajendra and Oh, Seong Yoon and Hagiwara, Yuki and Tan, Jen Hong and Adam, Muhammad},
  journal={Computers in Biology and Medicine},
  volume={94},
  pages={150--158},
  year={2018},
  publisher={Elsevier}
}

@article{xai_limitations,
  title={Explainability in medicine in an era of AI-based clinical decision support systems},
  author={Tonekaboni, Soroush and Joshi, Shalmali and McCradden, Melissa D and Goldenberg, Anna},
  journal={NPJ Digital Medicine},
  volume={2},
  pages={1--5},
  year={2019},
  publisher={Nature Publishing Group}
}

@inproceedings{lundberg2017unified,
  title={A unified approach to interpreting model predictions},
  author={Lundberg, Scott M. and Lee, Su-In},
  booktitle={Advances in Neural Information Processing Systems (NeurIPS)},
  volume={30},
  pages={4765--4774},
  year={2017}
}

@inproceedings{ribeiro2016should,
  title={"Why should I trust you?": Explaining the predictions of any classifier},
  author={Ribeiro, Marco Tulio and Singh, Sameer and Guestrin, Carlos},
  booktitle={Proceedings of the 22nd ACM SIGKDD International Conference on Knowledge Discovery and Data Mining (KDD)},
  pages={1135--1144},
  year={2016},
  organization={ACM}
}

@article{pumplun2021adoption,
  title={Adoption of machine learning systems for medical diagnostics in clinics: qualitative interview study},
  author={Pumplun, Luisa and Fecho, Mariska and Wahl, Nihal and Peters, Felix and Buxmann, Peter},
  journal={Journal of Medical Internet Research},
  volume={23},
  number={10},
  pages={e29301},
  year={2021},
  publisher={JMIR Publications Toronto, Canada}
}

@article{almansouri2024early,
  title={Early diagnosis of cardiovascular diseases in the era of artificial intelligence: An in-depth review},
  author={Almansouri, Naiela E and Awe, Mishael and Rajavelu, Selvambigay and Jahnavi, Kudapa and Shastry, Rohan and Hasan, Ali and Hasan, Hadi and Lakkimsetti, Mohit and AlAbbasi, Reem Khalid and Guti{\'e}rrez, Brian Criollo and others},
  journal={Cureus},
  volume={16},
  number={3},
  year={2024},
  publisher={Cureus}
}

@dataset{ribeiro_2021_4916206,
  author       = {Ribeiro, Antônio H. and
                  Paixao, Gabriela M.M. and
                  Lima, Emilly M. and
                  Horta Ribeiro, Manoel and
                  Pinto Filho, Marcelo M. and
                  Gomes, Paulo R. and
                  Oliveira, Derick M. and
                  Meira Jr, Wagner and
                  Schon, Thömas B and
                  Ribeiro, Antonio Luiz P.},
  title        = {CODE-15\%: a large scale annotated dataset of
                   12-lead ECGs
                  },
  month        = jun,
  year         = 2021,
  publisher    = {Zenodo},
  version      = {1.0.0},
  doi          = {10.5281/zenodo.4916206},
  url          = {https://doi.org/10.5281/zenodo.4916206},
}

@dataset{ribeiro_2020_3765780,
  author       = {Ribeiro, Antonio H and
                  Ribeiro, Manoel Horta and
                  Paixão, Gabriela M. and
                  Oliveira, Derick M. and
                  Gomes, Paulo R. and
                  Canazart, Jéssica A. and
                  Ferreira, Milton P. and
                  Andersson, Carl R. and
                  Macfarlane, Peter W. and
                  Meira Jr., Wagner and
                  Schön, Thomas B. and
                  Ribeiro, Antonio Luiz P.},
  title        = {CODE-test: An annotated 12-lead ECG dataset},
  month        = jan,
  year         = 2020,
  publisher    = {Zenodo},
  version      = {v1.0.3},
  doi          = {10.5281/zenodo.3765780},
  url          = {https://doi.org/10.5281/zenodo.3765780},
}

@article{targ2016resnet,
  title={Resnet in resnet: Generalizing residual architectures},
  author={Targ, Sasha and Almeida, Diogo and Lyman, Kevin},
  journal={arXiv preprint arXiv:1603.08029},
  year={2016}
}

@article{marey2024explainability,
  title={Explainability, transparency and black box challenges of AI in radiology: impact on patient care in cardiovascular radiology},
  author={Marey, Ahmed and Arjmand, Parisa and Alerab, Ameerh Dana Sabe and Eslami, Mohammad Javad and Saad, Abdelrahman M and Sanchez, Nicole and Umair, Muhammad},
  journal={Egyptian Journal of Radiology and Nuclear Medicine},
  volume={55},
  number={1},
  pages={183},
  year={2024},
  publisher={Springer}
}

@article{quinn2022three,
  title={The three ghosts of medical AI: Can the black-box present deliver?},
  author={Quinn, Thomas P and Jacobs, Stephan and Senadeera, Manisha and Le, Vuong and Coghlan, Simon},
  journal={Artificial intelligence in medicine},
  volume={124},
  pages={102158},
  year={2022},
  publisher={Elsevier}
}

@article{singh2022interpretation,
  title={Interpretation and classification of arrhythmia using deep convolutional network},
  author={Singh, Prateek and Sharma, Ambalika},
  journal={IEEE Transactions on Instrumentation and Measurement},
  volume={71},
  pages={1--12},
  year={2022},
  publisher={IEEE}
}

@incollection{aggarwal2022ecg,
  title={Ecg classification and analysis for heart disease prediction using xai-driven machine learning algorithms},
  author={Aggarwal, Ritu and Podder, Prajoy and Khamparia, Aditya},
  booktitle={Biomedical data analysis and processing using explainable (XAI) and responsive artificial intelligence (RAI)},
  pages={91--103},
  year={2022},
  publisher={Springer}
}

@article{sathi2024interpretable,
  title={An interpretable electrocardiogram-based model for predicting arrhythmia and ischemia in cardiovascular disease},
  author={Sathi, Tanjila Alam and Jany, Rafsan and Ela, Razia Zaman and Azad, AKM and Alyami, Salem Ali and Hossain, Md Azam and Hussain, Iqram},
  journal={Results in Engineering},
  volume={24},
  pages={103381},
  year={2024},
  publisher={Elsevier}
}

@misc{tensorflow,
  title        = {TensorFlow},
  howpublished = {\url{https://www.tensorflow.org}},
  note         = {Software library},
  year         = {2025}
}

@misc{pytorch,
  title        = {PyTorch},
  howpublished = {\url{https://pytorch.org}},
  note         = {Software library},
  year         = {2025}
}

@misc{scikitlearn,
  title        = {scikit-learn},
  howpublished = {\url{https://scikit-learn.org}},
  note         = {Software library},
  year         = {2025}
}

@article{budd2021survey,
  title={A survey on active learning and human-in-the-loop deep learning for medical image analysis},
  author={Budd, Samuel and Robinson, Emma C and Kainz, Bernhard},
  journal={Medical image analysis},
  volume={71},
  pages={102062},
  year={2021},
  publisher={Elsevier}
}

@article{konyushkova2017learning,
  title={Learning active learning from data},
  author={Konyushkova, Ksenia and Sznitman, Raphael and Fua, Pascal},
  journal={Advances in neural information processing systems},
  volume={30},
  year={2017}
}

@incollection{biswas2023active,
  title     = {Active Learning on Medical Image},
  author    = {Biswas, Angona and Abdullah Al, Nasim Md and Ali, Md Shahin
               and Hossain, Ismail and Ullah, Md Azim and Talukder, Sajedul},
  booktitle = {Data Driven Approaches on Medical Imaging},
  pages     = {51--67},
  year      = {2023},
  publisher = {Springer},
  address   = {Cham}
}

@article{das2023survey,
  title={A survey on EEG data analysis software},
  author={Das, Rupak Kumar and Martin, Anna and Zurales, Tom and Dowling, Dale and Khan, Arshia},
  journal={Sci},
  volume={5},
  number={2},
  pages={23},
  year={2023},
  publisher={MDPI}
}

@article{takei2015toward,
  title={Toward flexible and wearable human-interactive health-monitoring devices},
  author={Takei, Kuniharu and Honda, Wataru and Harada, Shingo and Arie, Takayuki and Akita, Seiji},
  journal={Advanced healthcare materials},
  volume={4},
  number={4},
  pages={487--500},
  year={2015},
  publisher={Wiley Online Library}
}

@inproceedings{de2020novel,
  title={A novel data collection framework for telemetry and anomaly detection in industrial iot systems},
  author={De Vita, Fabrizio and Bruneo, Dario and Das, Sajal K},
  booktitle={2020 IEEE/ACM fifth international conference on Internet-of-things design and implementation (IoTDI)},
  pages={245--251},
  year={2020},
  organization={IEEE}
}

@inproceedings{procopiuc2003bkd,
  title={Bkd-tree: A dynamic scalable kd-tree},
  author={Procopiuc, Octavian and Agarwal, Pankaj K and Arge, Lars and Vitter, Jeffrey Scott},
  booktitle={International symposium on spatial and temporal databases},
  pages={46--65},
  year={2003},
  organization={Springer}
}

@article{jacobson2004analyzing,
  title={Analyzing the performance of generalized hill climbing algorithms},
  author={Jacobson, Sheldon H and Y{\"u}cesan, Enver},
  journal={Journal of Heuristics},
  volume={10},
  number={4},
  pages={387--405},
  year={2004},
  publisher={Springer}
}
\end{document}